\documentclass[twocolumn]{svjour3}         

\smartqed 

\usepackage{graphicx}
\usepackage[pagebackref=false,breaklinks=true,letterpaper=true,colorlinks,bookmarks=false,citecolor=blue]{hyperref}
\usepackage{natbib}

\usepackage{dsfont}
\usepackage{multirow}

\usepackage{epsfig}
\usepackage{epstopdf}
\usepackage{color}

\usepackage{url}
\usepackage{array}
\usepackage{textcomp}
\usepackage{stfloats}
\usepackage{url}
\usepackage{verbatim}
\usepackage{graphicx}

\usepackage{amsmath,amsfonts}
\usepackage{algorithmic}
\usepackage{algorithm}
\usepackage{subcaption} 
\usepackage[table]{xcolor}
\definecolor{LightRed}{rgb}{1,0.7,0.7}

\begin{document}

\title{SwinTextSpotter v2: Towards Better Synergy for Scene Text Spotting}

\author{Mingxin Huang $ ^ 1$        
        \and
        Dezhi Peng $ ^ 1$             
        \and 
        Hongliang Li  $ ^ 1$          
        \and
        Zhenghao Peng  $ ^ 2$         
        \and
        Chongyu Liu $ ^ 1$            
        \and
        Dahua Lin   $ ^ 3$            
        \and
        Yuliang Liu  $ ^ {4*}$           
        \and
        Xiang Bai  $ ^ {4}$           
        \and
        Lianwen Jin  $ ^ {1}$         
}

\authorrunning{Mingxin Huang et al.}

\institute{ 
        $ ^1$ South China University of Technology, China \\
        $ ^2$ University of California, Los Angeles, USA \\
        $ ^3$ Chinese University of Hong Kong, China \\
        $ ^4$ Huazhong University of Science and Technology, China \\
}   

\date{Received: date / Accepted: date}

\maketitle

\begin{abstract}
End-to-end scene text spotting, which aims to read the text in natural images, has garnered significant attention in recent years. However, recent state-of-the-art methods usually incorporate detection and recognition simply by sharing the backbone, which does not directly take advantage of the feature interaction between the two tasks. In this paper, we propose a new end-to-end scene text spotting framework termed SwinTextSpotter v2, which seeks to find a better synergy between text detection and recognition. Specifically, we enhance the relationship between two tasks using novel Recognition Conversion and Recognition Alignment modules. Recognition Conversion explicitly guides text localization through recognition loss, while Recognition Alignment dynamically extracts text features for recognition through the detection predictions. This simple yet effective design results in a concise framework that requires neither an additional rectification module nor character-level annotations for the arbitrarily-shaped text. Furthermore, the parameters of the detector are greatly reduced without performance degradation by introducing a Box Selection Schedule. Qualitative and quantitative experiments demonstrate that SwinTextSpotter v2 achieves state-of-the-art performance on various multilingual (English, Chinese, and Vietnamese) benchmarks. The code will be available at \href{https://github.com/mxin262/SwinTextSpotterv2}{SwinTextSpotter v2}.
\keywords{Scene text spotting \and Synergy \and Text detection \and Text recognition \and Transformer}
\end{abstract}

\section{Introduction}\label{sec1}

Scene text spotting, which entails finding and recognizing text within natural images, has gained significant attention in recent years due to its practical applications in fields such as autonomous driving~\citep{9551780}, intelligent navigation~\citep{wang2015bridging,rong2016guided}, and key entities recognition~\citep{zhang2020trie,wang2021towards}, \textit{etc}. Despite recent advancements, text spotting remains a complex and unresolved issue due to the presence of diverse background noise and substantial variations in text shape, color, font, language, and layout.

Classical scene text spotting methods commonly view text spotting as two separate tasks that first find the location of the text and then transform the detected regions into character sequences by the recognizer~\citep{jaderberg2016reading,liao2018textboxes++,neumann2015real,gomez2017textproposals}. However, this pipeline has several drawbacks, including (1) error accumulation between text detection and recognition, \emph{i.e.}, the performance of text recognition is very sensitive to the results of text detection; 
(2) sub-optimal problem caused by optimizing two tasks separately; 
(3) intensive memory consumption and low inference efficiency.      

\begin{figure}[t!]
\centering
    
    \includegraphics[width=\linewidth]{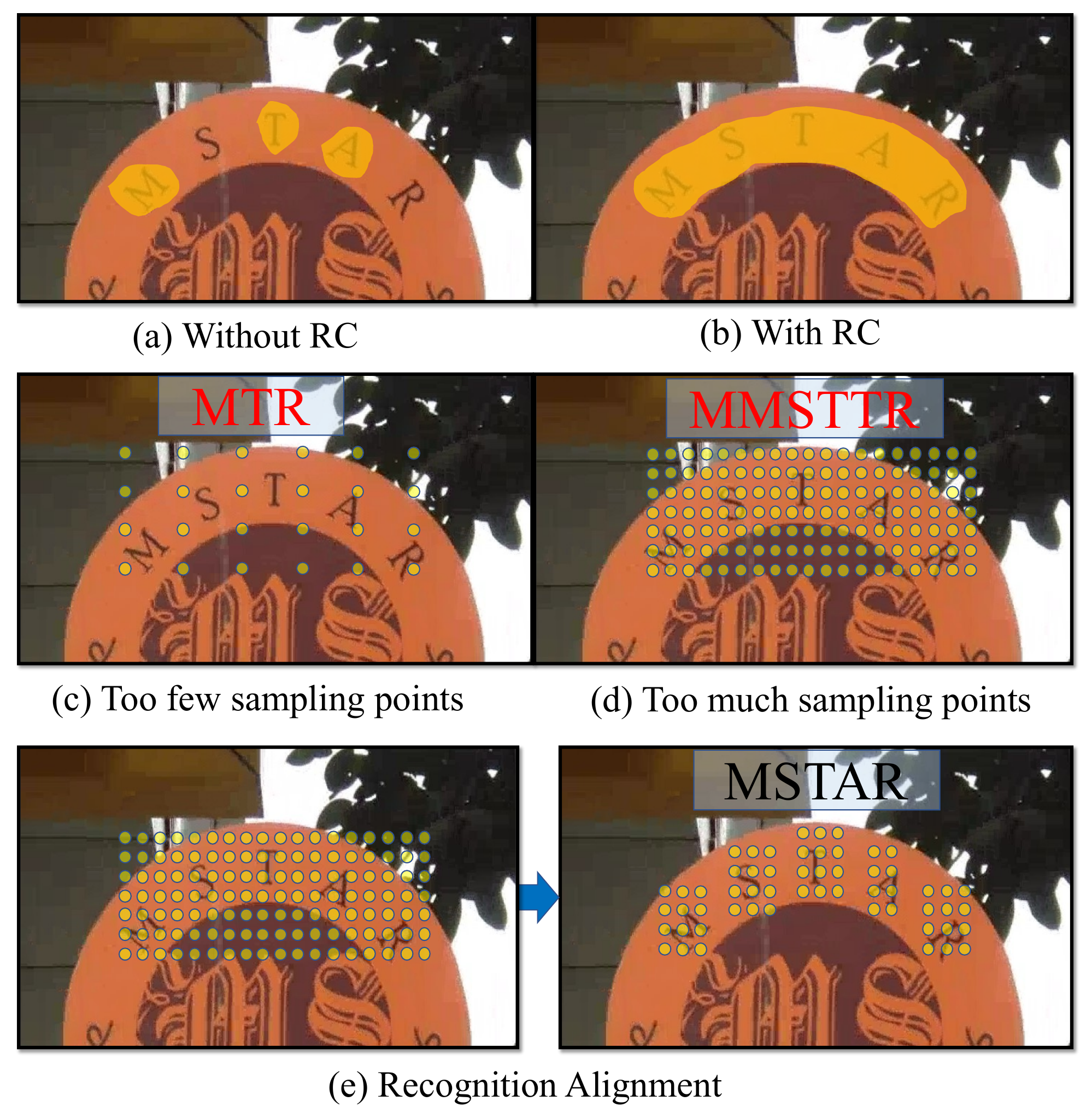}
    \caption{
    The effectiveness of Recognition Conversion (RC) and Recognition Alignment (RA). Previous connector-based methods usually struggle in the misalignment issue. Too few sampling points result in the failure to capture certain characters in the text region, leading to a loss of critical information. Too many sampling points produce a lot of background noise, thereby misleading the recognizer. Recognition Alignment can dynamically sample the text feature. 
        }\label{fig:Compared with RC and RA}
\end{figure}

In response to these challenges, there has been a recent trend toward~\citep{li2017towards,he2018end,liu2018fots,lyu2018mask,peng2022pagenet} integrating text detection and recognition into a single end-to-end system by sharing backbone. Through the end-to-end system, the recognizer can enhance the performance of the detector by reducing false positive predictions~\citep{li2017towards,liao2019mask}. 
In turn, even if the detection results are not completely accurate, the recognizer may still produce a correct text prediction due to its large receptive field on the feature map~\citep{liu2018fots,liao2019mask}. 
An additional advantage of an end-to-end system is that it is simpler to maintain and transfer to new domains, compared to a cascaded pipeline, which is tightly connected to the data and requires significant engineering efforts~\citep{liu2021abcnetv2,wang2021pan++}.

\begin{table}[t]
\centering
\caption{Influence of different density of sampling points. All the values are from ~\cite{liao2019mask} and ~\cite{liu2020abcnet}.}
\resizebox{\linewidth}{!}{
\begin{tabular}{l|c|c|c}
\hline
\multirow{2}*{Method} & \multirow{2}*{Sampling points} & ICDAR2015 & Total-Text                     \\ \cline{3-4} 
           &      & G   & None \\ \hline
Mask TextSpotter        & (8,32)           & 71.3                    & \multicolumn{1}{c}{-}  \\ 
Mask TextSpotter        & (32,32)           & 73.2                    & \multicolumn{1}{c}{-}  \\ 
Mask TextSpotter        & (16,64)           & \textbf{73.5}                    & \multicolumn{1}{c}{-}  \\ 
Mask TextSpotter        & (16,128)           & 71.6                    & \multicolumn{1}{c}{-}  \\ 
\hline
ABCNet        & (6,32)           & -                    &  59.6   \\ 
ABCNet        & (7,32)           & -                    & \textbf{61.9}  \\ 
ABCNet        & (14,64)           & -                    & 58.1  \\ 
ABCNet        & (21,96)           & -                   & 54.8  \\ 
\hline
\end{tabular}}
\label{tab:samplepoint}
\end{table}

However, most existing end-to-end scene text spotting systems struggle in three primary aspects~\citep{lyu2018mask,feng2019textdragon,qin2019towards,qiao2020text,wang2020all,liao2020mask,liu2020abcnet}. 
First, if the detection relies solely on the visual information, the detector may be easily disturbed by background noise and generate inconsistent detection results, as shown in Fig.~\ref{fig:Compared with RC and RA}(a).
Interaction between texts within the same image is a critical factor in eliminating the impact of background noise as different characters of the same word may exhibit strong similarities in terms of their backgrounds and styles. 
One alternative is to utilize a Transformer~\citep{vaswani2017attention} to capture rich interactions between text instances. For example, ~\cite{yu2020towards} use Transformer to make texts interact with each other at a semantic level. ~\cite{fang2021read} and ~\cite{wang2021two} further adopt Transformer to model the visual relationship between texts.
Second, despite sharing a backbone, the interaction between detection and recognition is insufficient, as neither the detector is optimized by the recognition loss nor does the recognizer make use of the detection features. 
In order to jointly enhance both detection and recognition, Mask TextSpotter~\citep{liao2020mask} simultaneously optimizes both tasks within a single branch by introducing a character segmentation map;
ABCNet v2~\citep{liu2021abcnetv2} proposes an Adaptive End-to-End Training (AET) strategy that utilizes the detection results to extract recognition features, as opposed to relying solely on ground truths;
ARTS ~\citep{zhong2021arts} improves the performance of the end-to-end text spotting by back-propagating the loss from the recognition branch to the detection branch using a differentiable Spatial Transform Network (STN)~\citep{2015Spatial}. 
However, these three methods assume the detector proposes detection predictions structurally, \textit{e.g.} in the reading order. The overall performance of the text spotting is thereafter bounded by the detector. Thirdly, we have observed that connector-based text spotters~\citep{liao2019mask,liu2020abcnet} are sensitive to the sampling points in the Region of Interest (RoI) operation, as presented in Table~\ref{tab:samplepoint}. Insufficient sampling points may result in the failure to capture certain characters in the text region, leading to a loss of critical information. Conversely, an excessive number of sampling points may produce a lot of background noise, thereby misleading the recognizer. Previous studies have regarded this phenomenon as a problem of sampling point selection~\citep{liao2019mask,liu2020abcnet}. In this paper, we further formulate it as a misalignment issue. Fig.~\ref{fig:Compared with RC and RA}(c) provides an example of insufficient sampling points, where the sampling points fail to capture the characters `T' and `S', requiring the recognition of these two characters through the receptive field of adjacent features. This situation poses a significant challenge for backbone feature extraction. Fig.~\ref{fig:Compared with RC and RA}(d) provides an example of too many sampling points, where the sampling points introduce too much background noise from the gap between different characters. Although previous methods achieve promising results, they require carefully selecting the optimal position for sampling features to address the misalignment issue.

In this paper, we propose \textit{SwinTextSpotter v2}, an end-to-end trainable Transformer-based framework, stepping toward better synergy between text detection and recognition.
To better distinguish the densely scattered text instances in crowded scenes, we introduce Transformer and a two-level self-attention mechanism in SwinTextSpotter v2, stimulating the interactions between the text instances.
To address the challenge in arbitrary-shape scene text spotting, inspired by ~\cite{sun2021sparse,hu2021istr}, we regard the text detection task as a set-prediction problem and thus adopt a query-based text detector.
We further propose \textit{Recognition Conversion (RC)} which implicitly guides the recognition head through incorporating the detection features and \textit{Recognition Alignment (RA)} which dynamically samples the
text feature to solve the misalignment problem. RC can back-propagate recognition information to the detector and suppress the background noise in the recognition features, leading to the joint optimization of the detector and recognizer. RA, on the other hand, dynamically samples the text feature to connect the detector and recognizer, ensuring that all characters in the text region are sampled. Empowered by the proposed RC and RA, SwinTextSpotter v2 has a concise framework without the character-level annotation and rectification module used in previous works to improve the recognizer. To simplify the detector and reduce the parameters, we adopt a straightforward and efficient Box Selection Schedule method to generate high-quality proposal boxes for the detector, thus reducing the need for the refinement stage in the query-based detector. SwinTextSpotter v2 has superior performance in both detection and recognition. As illustrated in Fig.~\ref{fig:Compared with RC and RA}(b), the detector of SwinTextSpotter v2 can accurately localize difficult samples with RC. The recognizer of SwinTextSpotter v2 can accurately extract the content of text instance with RA, as compared to Fig. \ref{fig:Compared with RC and RA}(e).

In this paper, the main improvement over the conference version~\citep{huang2022swintextspotter} lies in the restructuring of the architecture between the text detector and recognizer. In our earlier work, we develop a text detector with six refinement stages to enable effective detection. However, such a design results in an excessive number of parameters. To address this issue, we attempt to reduce the number of stages in the detector. However, doing so could result in a great decrease in performance. Recent studies, such as~\citep{zhu2020deformable,zhang2022dino,jia2022detrs,ouyangzhang2022nms,zhang2022expected}, have shown that query-based detector can significantly benefit from high-quality proposal boxes. In this paper, we utilize a Box Selection Schedule to generate high-quality proposal boxes for the query-based detector, enabling the detector to maintain optimal performance with only three refinement stages. Furthermore, we have observed that even with accurate detection results, the recognizer still struggles to recognize some text instances that exhibit simple shapes and font styles. We identify this issue as a misalignment issue in text spotting. The misalignment is caused by an incorrect sampling location of text features in the RoI operation, which significantly impacts text spotting performance. To address this issue, we propose a \textit{Recognition Alignment} to enable dynamic sampling of text features, thereby enhancing the overall performance of the text spotting system.

We conduct extensive experiments on seven benchmarks, including multi-oriented dataset RoIC13~\citep{liao2020mask} and ICDAR 2015~\citep{karatzas2015icdar}, multilingual dataset ReCTS (Chinese)~\citep{zhang2019icdar} and VinText (Vietnamese)~\citep{m_Nguyen-etal-CVPR21}, and arbitrarily-shaped dataset Total-Text~\citep{ch2020total}, SCUT-CTW1500~\citep{liu2019curved} and Inverse-Text~\citep{ye2022dptext}. The results demonstrate the superior performance of the SwinTextSpotter v2.

The main contributions of this paper are summarized as follows.
\begin{itemize}

\item We propose a \textit{Recognition Conversion} to enhance the synergy of text detection and recognition.
\item We observe a misalignment issue in the feature sampling for the recognition of text spotting. To solve this issue, we develop a \textit{Recognition Alignment} to dynamically sample the recognition features through the detection predictions.
\item SwinTextSpotter v2 is a concise framework that does not require character-level annotation as well as a specifically designed rectification module for recognizing arbitrarily-shaped text.
\item We greatly reduce the parameters by approximately 50\% compared to the conference version while achieving better performance by introducing a Box Selection Schedule.
\item SwinTextSpotter v2 achieves state-of-the-art performance on multiple public scene text benchmarks. 
\end{itemize}

\begin{figure*}[thp]
    \centering
    \includegraphics[width=\textwidth]{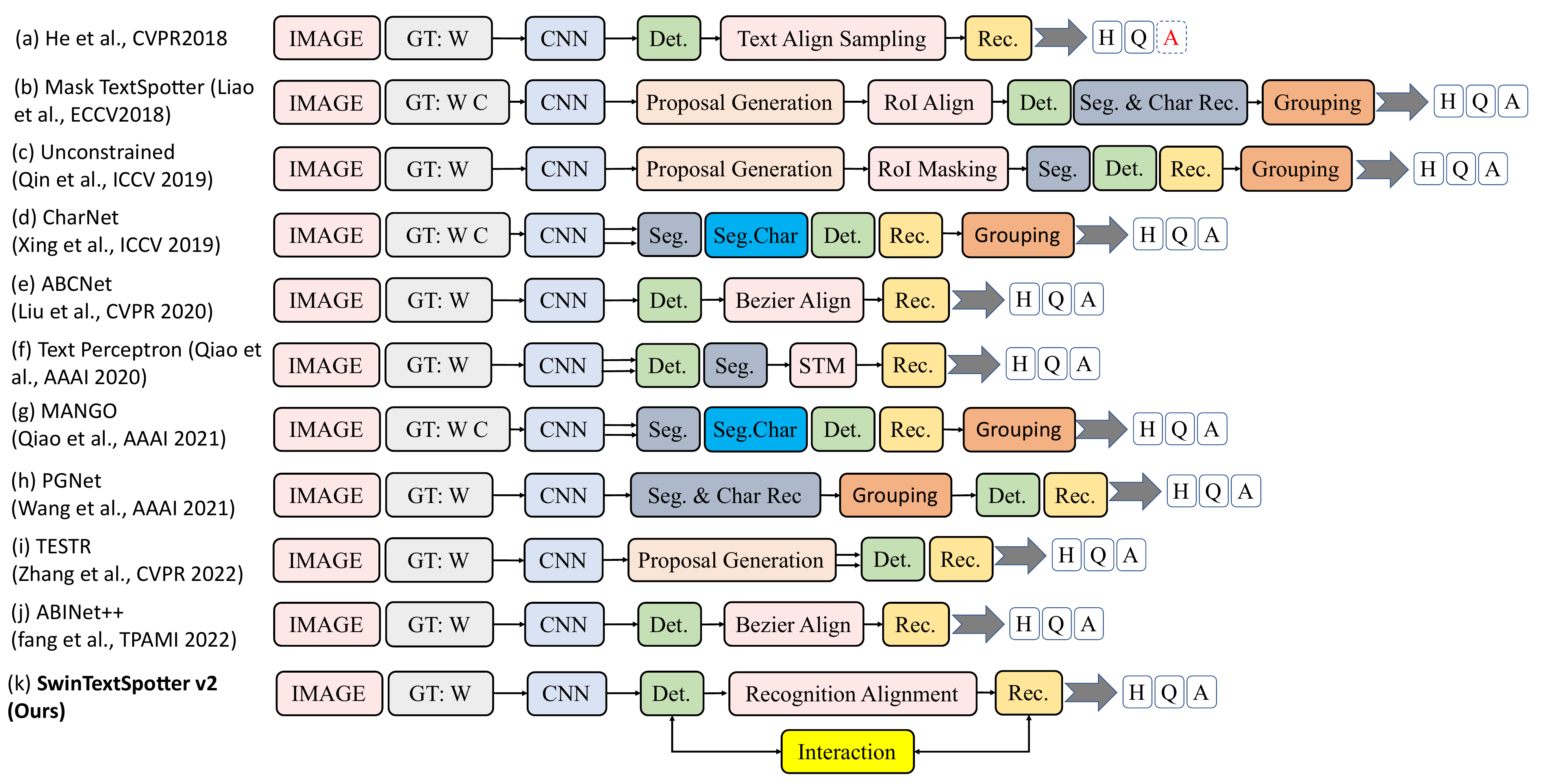}
    \caption{
        Overview of several end-to-end scene text spotting methods that correspond to our research objectives. Within the ground-truth (GT) box, the letters 'W' and 'C' indicate word-level and character-level annotations, respectively. The letters 'H', 'Q', and 'A' signify the method's ability to detect horizontal, quadrilateral, and arbitrarily-shaped text, respectively. Conversely, the dashed box denotes the shape of the text that the method cannot detect. Figure style from~~\citep{liu2020abcnet,wang2021pgnet,zhang2022text}
    }
    \label{fig:review}

\end{figure*}

\section{Related Work}
Although previous text spotting methods have achieved remarkable progress, the synergy between text detection and recognition is still under exploration. Fig.~\ref{fig:review} is an overview of exemplary works.

\textbf{Separate Scene Text Spotting}.
Traditional text spotting approaches typically divide the task into two separate systems, with text detection and text recognition being performed independently of one another. This approach neglects the potential synergy that could exist between the two processes. ~\cite{wang2011end} develop a frame to locate the text by a sliding-window-based detector and use a character classifier to classify each character. ~\cite{bissacco2013photoocr} build a characters classification system by combining DNN and HOG features. ~\cite{liao2018textboxes++} incorporate the single-shot text detector~\citep{liao2017textboxes} to locate the text and attach a text recognizer~\citep{shi2016end} to recognize the text content.

\textbf{End-to-End Text Spotting}.
Recently, many researchers pay more attention to developing end-to-end spotters to obtain an ideal performance. ~\cite{li2017towards} propose an end-to-end trainable scene text spotting framework to combine text detection and 
 recognition by sharing the backbone. FOTS~\citep{liu2018fots} uses a one-stage detector to generate rotated boxes and adopts RoIRotate to sample the oriented text feature into horizontal grids for connecting the text detection and recognition. ~\cite{he2018end} propose a similar framework using an attention-based recognizer.

For the task of arbitrarily-shaped scene text spotting, 
Mask TextSpotter series~\citep{lyu2018mask,liao2019mask,liao2020mask} solve the problem without explicit rectification by using the character segmentation branch to improve the performance of the recognizer.
TextDragon~\citep{feng2019textdragon} combines the two tasks by RoISlide, a technique that transforms the predicted segments of text instances into horizontal features. \cite{wang2020all} adopt Thin-Plate-Spline~\citep{bookstein1989principal} transformation to rectify the features. ABCNet~\citep{liu2020abcnet} and its improved version ABCNet v2~\citep{liu2021abcnetv2} use the BezierAlign to transform the arbitrary-shape texts into regular ones. These methods achieve great progress by using rectification modules to unify text detection and recognition into end-to-end trainable systems. 
~\cite{qin2019towards} propose RoI Masking to extract the feature for arbitrarily-shaped text recognition. Similar to \cite{qin2019towards}, PAN++~\citep{wang2021pan++} is based on a faster detector~\citep{wang2019efficient}. AE TextSpotter~\citep{wang2020ae} uses the results of recognition to guide detection through a language model. ABINet++~\citep{fangabinet++} introduces an autonomous, bidirectional, and iterative language model into the recognizer to utilize linguistic knowledge. Though achieve significant improvement in the performance of text spotting by sharing backbone, the aforementioned methods neither back-propagate recognition loss to the detector nor use detection features in the recognizer. The detector and the recognizer thus are still relatively independent of each other without joint optimization.
Recently, ~\cite{zhong2021arts} propose ARTS which passes the gradient of recognition loss to the detector using Spatial Transform Network (STN)~\citep{2015Spatial}, demonstrating the power of synergy between the detection and recognition in text spotting.

\textbf{Detection Transformer in Text Spotting}.
Inspired by the detection transformer~\citep{carion2020end,zhu2020deformable}, recent methods attempt to address text spotting using Transformer structures. For instance, TESTR~\citep{zhang2022text} develops a dual decoder to locate and recognize text separately to get rid of the complicated post-processing. TTS~\citep{kittenplon2022towards} adds an RNN recognition head in the Deformable DETR~\citep{zhu2020deformable} and further proposes a weak supervision approach. More recently, inspired by an auto-regressive algorithm~\citep{chen2021pix2seq}, SPTS~\citep{peng2022spts} formulates text detection as a point detection problem. Although the aforementioned methods demonstrate the efficacy of utilizing the interaction between text instances in Transformer structures, the synergy between text detection and recognition is still under-explored.

\begin{figure*}[htp]
    \centering
    \includegraphics[width=\textwidth]{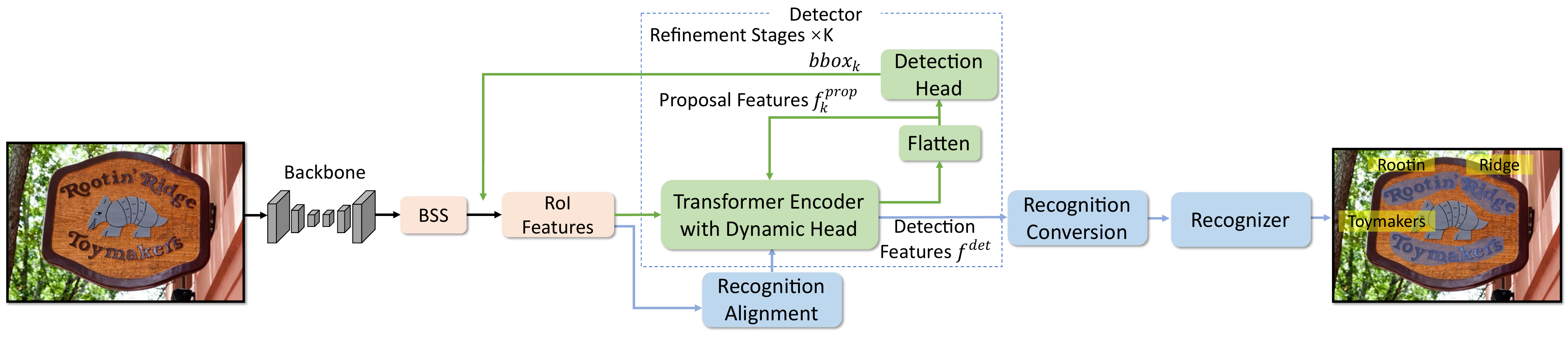}
    \caption{
        The framework of the proposed SwinTextSpotter v2. The gray arrows denote the feature extraction from images. The green arrows and blue arrows represent the detection stage and the recognition stage, respectively. The outputs of the detection head are refined in K stages. The output detection in the $K^{th}$ stage serves as the input to the recognition stage.
    }
    \label{fig:frame}

\end{figure*}

\section{Methodology}
\label{sec:method}

The overall architecture of SwinTextSpotter v2 is presented in Fig.~\ref{fig:frame}, which consists of four components: 
(1) a backbone based on Swin-Transformer~\citep{liu2021Swin};
(2) a box selection schedule to generate high-quality proposal boxes; 
(3) a query-based text detector; 
(4) a \textit{Recognition Alignment} and \textit{Recognition Conversion} to bridge the text detector and recognizer; and 
(5) an attention-based recognizer. 

As illustrated in the green arrows of Fig.~\ref{fig:frame}, in the first stage of detection, we obtain the high-quality proposal boxes $bbox_0$ through the Box Selection Schedule, and randomly initialize trainable parameters to be proposal features $f_0^{prop}$. We add the global information into the proposal features by extracting the image features and adding them into $f_0^{prop}$. Then, the RoI features are obtained by using the RoI align~\citep{he2017mask} with the proposal boxes $bbox_0$. The RoI features and $f_0^{prop}$ are fused in the Transformer encoder with dynamic head. The new proposal features $f_1^{prop}$ are formed by flattening the output of the Transformer encoder and the new proposal features $f_1^{prop}$ are fed into the detection head to output the detection result. The input of later $k^{th}$ detection stage is the detection results $bbox_{k-1}$ and proposal feature $f_{k-1}^{prop}$. The boxes $bbox_{k-1}$ and the proposal feature $f_{k-1}^{prop}$ are updated layer by layer in the detector. Compared to the conference version that utilizes six stages for refinement, we have reduced the refinement stage to three to decrease the detector's parameters. However, this reduction will lead to a decrease in performance. To mitigate this issue, we further develop a Box Selection Schedule (BSS) to prevent performance degradation. More details of the Box Selection Schedule and the detector are explained in Section \ref{sec:Box Selection Schedule} and \ref{sec:A Query Based Detector}, respectively.

The resolution of the recognition and detection are different, so we use a higher rate of resolution in the recognition stage (blue arrows). The recognition features are obtained by using the final detection stage output box $bbox_K$. Then the recognition features are sent into \textit{Recognition Alignment} to dynamically align the text features. In order to keep the resolution of features consistent with the detector when fused with proposal features, we down-sample the RoI features in the RA to get three feature maps of descending sizes in \textit{Recognition Alignment}, denoting by $\{a_1,a_2,a_3\}$. Then we obtain detection features $f^{det}$ by fusing the smallest $a_3$ and the proposal features $f_K^{prop}$. The detection features $f^{det}$ in the recognition stage contain all previous detection information. Finally the $\{a_1,a_2,a_3\}$ and the detection features $f^{det}$ are sent into \textit{Recognition Conversion} and recognizer for generating the recognition result.
More details of \textit{Recognition Alignment}, \textit{Recognition Conversion} and recognizer are explained in Section \ref{sec:Recognition Alignment}, Section \ref{sec:Recognition Conversion} and Section \ref{sec:recognizer}, respectively.

\begin{figure}[t!]
    \centering
    \includegraphics[width=\linewidth]{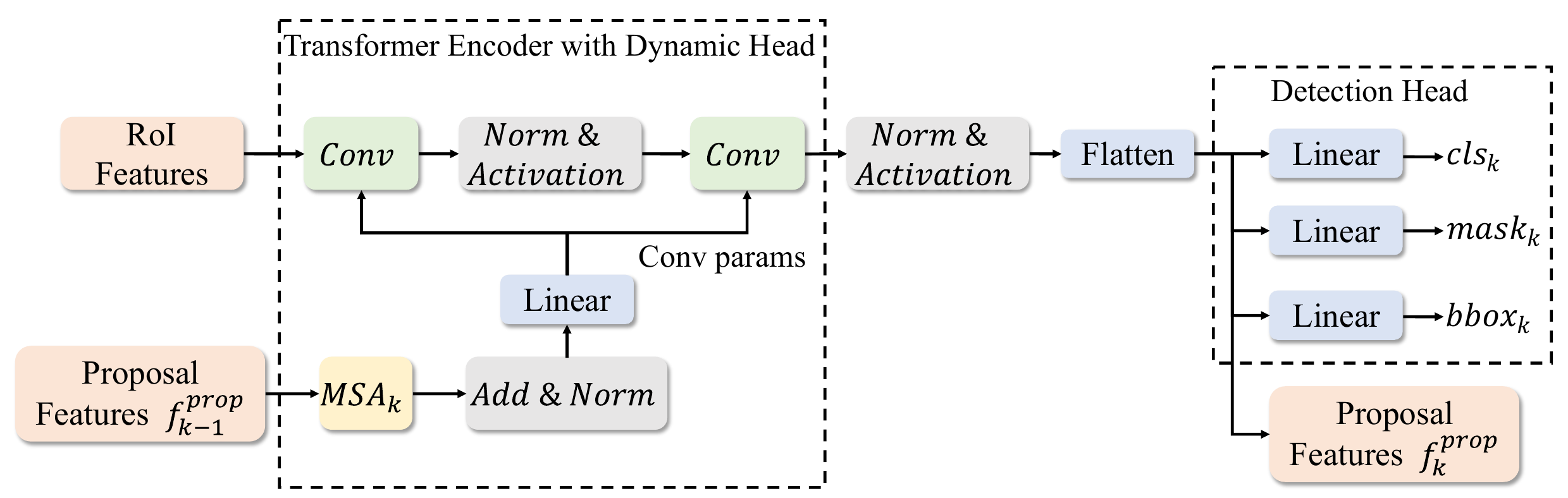}
    \caption{Illustration of $k^{th}$ stage in detection. $f_{k-1}^{prop}$ is the proposal features output by previous stage. 
    $MSA_{k}$ refers to the multi-head-attention in $k^{th}$ stage. $f_{k}^{prop}$ will serve as the input to next stage.
    }
    \label{fig:detector}
\end{figure}

\subsection{Box Selection Schedule}
\label{sec:Box Selection Schedule}
The training convergence of query-based detectors can be slow due to their sparse design. To address this challenge, various solutions have been proposed in recent works such as~\citep{zhu2020deformable,meng2021conditional,liu2022dabdetr,li2022dn,ouyangzhang2022nms,zhang2022expected}. One of these solutions involves generating high-quality proposal boxes for the query-based detector, which allows for more precise pooling of text instance features.

In the conference version, we randomly initialize proposal boxes $bbox_0$. In this study, we develop a Box Selection Schedule (BSS) to generate high-quality proposal boxes. The BSS consists of three convolution layers implementing classification and regression in two subtasks. First, we present a series of anchor boxes in the features map following the previous works~\citep{ren2015faster,he2017mask}. Then we adopt the $P_3$ to $P_6$ features, output from the Feature Pyramid Network (FPN)~\citep{lin2017feature}, to generate the proposal boxes. $P_l$ represents the $2^{(l-1)}$ downsampling level relative to the original image. To ensure the quality of the proposal boxes, every feature within the multi-scale feature maps generated by the Feature Pyramid Network (FPN) is utilized to produce a classification score and a corresponding proposal box. This step aims at achieving a high recall rate, ensuring that most of the text instances of interest are captured, even if it means generating a large number of low-quality proposals. Then, we employ a series of criteria to select the high-quality proposal boxes: 1) Non-Maximum Suppression (NMS): To eliminate redundant overlapping boxes, we apply NMS with an Intersection over Union (IoU) threshold. This ensures that for any given text instance, there is only one representative proposal box, thus reducing false positives and improving the overall quality of the selected boxes. 2) Top-K Selection: After applying NMS, we select the top K proposal boxes according to their classification scores, ensuring that the highest quality proposals are chosen for further processing. The top K proposal boxes are then utilized as the initial $bbox_0$ for the first stage of the query-based detector.

We utilize two types of loss to train the BSS, including classification and regression loss. The classification ${L}_{{cls}}$ loss is calculated using binary logistic loss, distinguishing between text and non-text regions. The regression loss $L_{{reg}}$ is the smooth $L_1$ loss~\citep{ren2015faster}. The loss function is defined as:
\begin{equation}
L_{{B\!S\!S}}=\frac{1}{N_{cls}} \sum_{i=1}^{N_{cls}} {L}_{{cls}} (p_i,p_i^\star)  + \frac{1}{N_{reg}}
\sum_{i=1}^{N_{reg}} L_{{reg}} (r_i,r_i^\star),
\end{equation}
where $N_{cls}$ is the number of anchor boxes in a mini-batch and $N_{reg}$ is the number of positive anchor boxes in this batch. An anchor box is classified as positive if its Intersection-over-Union (IoU) with any ground truth box exceeds 0.7, and as negative if the IoU is below 0.3. $p_i$ stands for the predicted probability of anchor $i$ being text and $p_i^\star$
is the associated ground-truth label. $r_i$ represents the predicted bounding box for anchor $i$,
and $r_i^\star$ is the corresponding ground-truth bounding box.

The Box Selection Schedule enhances the feature learning in the backbone and generates high-quality proposal boxes for the query-based detector. By employing BSS, we can utilize three refinement stages for the detector while achieving better performance than the conference version which requires six stages, thereby simplifying the detector and reducing the number of parameters.

\subsection{Query Based Detector}
\label{sec:A Query Based Detector}
We use a query-based detector to detect the text. Based on Sparse R-CNN ~\citep{sun2021sparse}, the query-based detector is built on ISTR~\citep{hu2021istr} which treats detection as a set-prediction problem. Our detector uses a set of learnable proposal features to represent high-level semantic vectors of text, and the proposal boxes generated by the BSS to help the proposal features aggregate the text features. The detector is empirically designed to have three refinement stages. With the Transformer encoder with dynamic head, the latter stages can access the information in former stages stored in the proposal features~\citep{jia2016dynamic,tian2020conditional,sun2021sparse}.
Through multiple stages of refinement, the detector can be applied to text at any scale.

The architecture of the detection head in $k^{th}$ stage is illustrated in Fig.~\ref{fig:detector}. The proposal features in $k-1$ stage is represented by $f_{k-1}^{prop} \in \mathbb R^{N, d}$.
At stage $k$, the proposal features $f_{k-1}^{prop}$ produced in the previous stage is fed into a self-attention module~\citep{vaswani2017attention} $MSA_{k}$ to model the relationships and generate two sets of convolutional parameters.
The detection information from previous stages is embedded into the convolutions which are used to encode RoI features. Then, the RoI features are extracted by $bbox_{k-1}$, which is the detection result from the previous stage, using RoIAlign~\citep{he2017mask}. The output features of the convolutions are then fed into a linear projection layer to produce $f_{k}^{prop}$ for the next stage. $f_{k}^{prop}$ is subsequently fed into the prediction head to generate $bbox_{k}$ and $mask_{k}$. To reduce computation, the 2D mask is transformed into a 1D mask vector by the Principal Component Analysis~\citep{wold1987principal} so the $mask_{k}$ is a one-dimensional vector.

When $k=1$, the $bbox_{0}$ is generated from the BSS, and $f_{0}^{prop}$ is randomly initialized parameters, which is the input of the first stage. 
During training, these parameters are updated via back-propagation and learn the inductive bias of the high-level semantic features of the text.

We view the text detection task as a set-prediction problem. Formally, we use the bipartite match to match the predictions and ground truths~~\citep{carion2020end,stewart2016end,sun2021sparse,hu2021istr}. The matching cost becomes:
\begin{equation}
L_{match} = \lambda_{cls} \cdot L_{cls} + \lambda_{L1} \cdot L_{L1} + \lambda_{giou} \cdot L_{giou} + \lambda_{mask} \cdot L_{mask},\label{XX}
\end{equation}
where $\lambda$ is the hyper-parameter used to balance the loss. $L_{cls}$ is the focal loss~\citep{lin2017focal}. 
The losses for regressing the bounding boxes are $L_1$ loss $L_{L1}$ and generalized IoU loss $L_{giou}$~\citep{rezatofighi2019generalized}. 
We compute the mask loss $L_{mask}$ following~\citep{hu2021istr}, which calculates the cosine similarity between the prediction mask and ground truth. The detection loss is similar to the matching cost but we use the $L_2$ loss and dice loss~\citep{milletari2016v} to replace the cosine similarity as in~\citep{hu2021istr}.

\begin{figure}[t!]
\centering
   
    \begin{minipage}{\linewidth}
        \includegraphics[width=8.0cm, height=6.6cm]{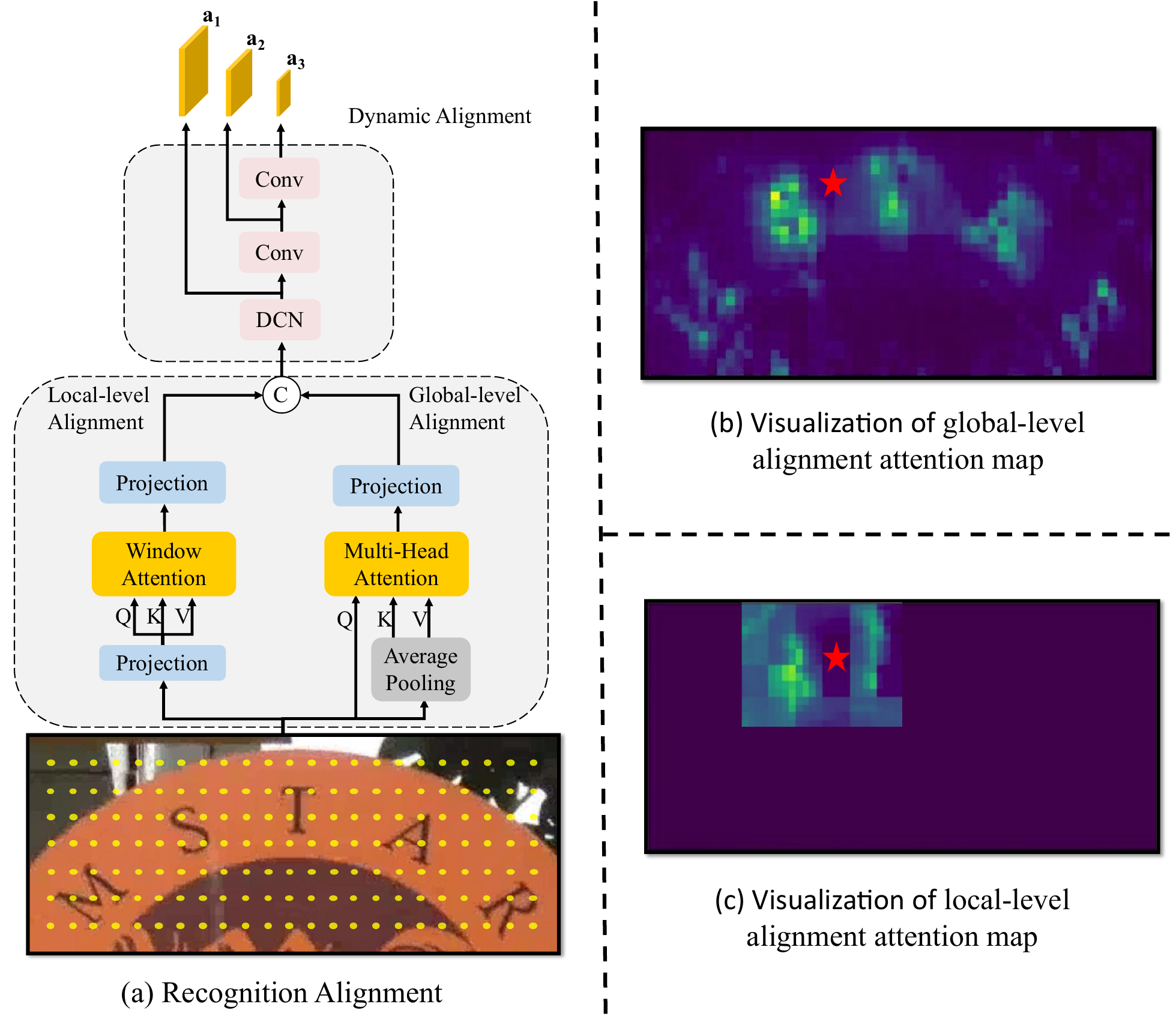}
        \label{fig:Recognition Alignment}
    \end{minipage}
    \caption{The architecture of Recognition Alignment and visualizations of global- and local-level alignment attention maps. The red star indicates the query point.
    }\label{fig:recognition alignment}
\end{figure}

\subsection{Recognition Alignment}
\label{sec:Recognition Alignment}

The details of \textit{Recognition Alignment} (RA) are illustrated in Fig.~\ref{fig:recognition alignment}(a). We use a single text instance as an example. Firstly, we use RoI Align~\citep{he2017mask} to extract a feature map $F_h \in \mathbb R^{H \times W \times C}$ from the detected bounding boxes. To accurately capture all characters within these boxes, we adopt a high-density sampling strategy that leverages RoI Align with densely sampled points ($64 \times 128$ points per RoI), outperforming conventional methods such as ABCNet ($7 \times 32$ points per RoI) and MaskTextSpotter v3 ($32 \times 32$ points per RoI). This strategy samples the features within each region of interest at a much finer granularity. Subsequently, these extracted features are fed into the RA module for alignment. Inspired by the LITv2~\citep{pan2022hilo}, we develop a two-level attention alignment structure. The features are fed into two branches. One is the local-level alignment. In this branch, we adopt the windows attention~\citep{liu2021Swin} to align the features, which aligns the text details locally. The local-level alignment can be expressed as
\begin{eqnarray}
Q = XW_q, K = XW_k, V = XW_v, \\
X^{l}=Proj(W-MSA(Q, K, V)),
\end{eqnarray}
X is the input feature. $W-MSA$ is the windows attention operator. $Proj$ is a projection layer. $X^{l}$ is the output of the low-level alignment. The projection layer $Proj$ merges the outputs from different ``heads" of windows attention operator, thereby integrating the information from various perspectives to form a richer and more comprehensive representation.

The other is the global-level alignment. Inspired by the PVT~\citep{wang2021pyramid}, in this branch, we use the average pooling in the input features as the keys $K \in \mathbb R^{HW/{S^2} \times C}$ and values $V \in \mathbb R^{HW/{S^2} \times C}$. S is the window size in average pooling. The features map $F_h$ as the queries. Then the standard attention is used to align the global-level semantic feature globally. The global-level alignment can be formulated as
\begin{eqnarray}
Q^{'} = XW_q^{'}, K^{'} = AP(X)W_k^{'}, V^{'} = AP(X)W_v^{'},
\end{eqnarray}

\begin{eqnarray}
X^{h} = Proj(MSA(Q^{'}, K^{'}, V^{'})),
\end{eqnarray}
X is the input feature. $Proj$ is a linear layer. $MSA$ is the multi-head attention operator. $X^{h}$ is the output of the high-level alignment. $AP$ is the average pooling operation.

Following previous works~\citep{carion2020end,zhu2020deformable}, we employ six rounds of the two-level alignment process. After this process, all features within the detection bounding box are aligned with the text. Specifically, for the same features, the global-level alignment branch globally aligns all characters in the text region, as presented in Fig.~\ref{fig:recognition alignment}(b), while the local-level alignment branch locally aligns characters adjacent to it, as presented in Fig.~\ref{fig:recognition alignment}(c). Then, we remove redundant features through a dynamic alignment process, which leverages deformable convolution's dynamic feature selection capability~\citep{dai2017deformable}. Specifically, the alignment consists of a deformable convolution layer followed by two standard convolutional layers: 1. Given \( F_h \in \mathbb{R}^{H \times W \times C} \), deformable convolution performs downsampling to adaptively remove redundant features, obtaining \( a_1 \in \mathbb{R}^{H/2 \times W/2 \times C} \). 2. Two cascaded convolutional layers further downsample \( a_1 \): the first produces \( a_2 \in \mathbb{R}^{H/4 \times W/4 \times C} \), and the second yields \( a_3 \in \mathbb{R}^{H/8 \times W/8 \times C} \). These features are then passed to the Recognition Conversion Module for further processing.

The dynamic sampling process in RA is learnable. We employ the recognition loss as the training signal to guide the dynamic sampling process. During training, to minimize the recognition loss, RA learns to dynamically select text features that are most beneficial for the recognizer. This leads to a more precise feature learning of the text instance and thus, an improvement in text spotting performance. In this manner, the Recognition Alignment alleviates the misalignment problem to enhance the overall performance of the text spotting system.

\subsection{Recognition Conversion}
\label{sec:Recognition Conversion}

\begin{figure}[t!]
\centering
    \includegraphics[width=\linewidth]{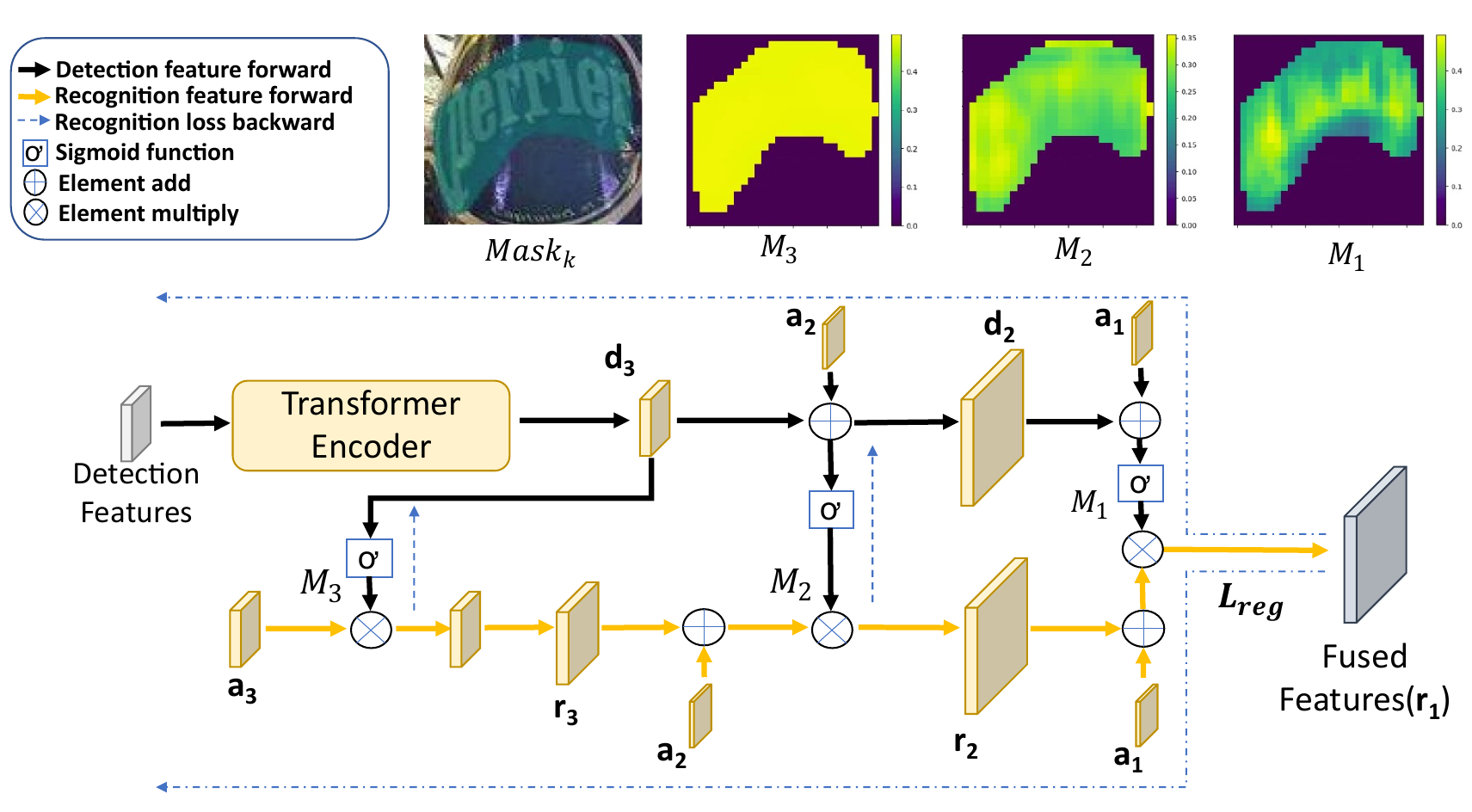}
    \caption{
        Detailed structure of \textit{Recognition Conversion}. In the masks $M_i = \phi{(d_i)}, i=1,2,3.$, the brighter the color, the higher the value.
    }
    \label{fig:recognition conversion}
\end{figure}

To better coordinate the detection and recognition, we propose \textit{Recognition Conversion (RC)} to spatially inject the features from the detection head into the recognition stage, as illustrated in Fig.~\ref{fig:recognition conversion}. 
The RC consists of the Transformer encoder ~\citep{vaswani2017attention} and four up-sampling structures. The input of RC are the detection features $f^{det}$ and three down-sampling features $\{a_3,a_2,a_1\}$.

The detection features are sent to the Transformer encoder $TrE()$, enabling the information of previous detection stages to further fuse with $a_3$. 
Then through a stack of up-sampling operation $E_u()$ and Sigmoid function $\phi{()}$, three masks $\{M_3,M_2,M_1\}$ for text regions are generated:
\begin{eqnarray}
d_3 = & TrE(f^{det}), \\
d_2 = & (E_u(d_3) + a_2), \\
d_1 = & (E_u(d_2) + a_1), \\
M_i = & \phi{(d_i)}, i=1,2,3.
\end{eqnarray}

With the masks $\{M_3, M_2, M_1\}$ and the input features $\{a_3,a_2,a_1\}$, we further integrate these features effectively under the following pipeline:
\begin{eqnarray}
r_3 = & M_3 \cdot a_3, \\
r_2 = & M_2 \cdot (E_u(r_3)+a_2), \\
r_1 = & M_1 \cdot (E_u(r_2)+a_1),
\end{eqnarray}

\noindent where $\{r_3, r_2, r_1\}$ denote the recognition features. The $r_1$ is the fused features in Fig.~\ref{fig:recognition conversion}, which is finally sent to the recognizer at the highest resolution. 
As shown in the blue dashed lines in Fig.~\ref{fig:recognition conversion},
the gradient of the recognition loss $L_{reg}$ can be back-propagated to the detection features, enabling RC to implicitly improve the detection head through the recognition supervision.

Generally, to suppress the background, the fused features will be multiplied by a $Mask_{K}$ predicted by the detection head (with the supervision of $L_{mask}$).
However, the background noise still remains in the feature maps as the detection box is not tight enough.
Such an issue can be alleviated by the proposed RC, since RC uses the detection features to generate tight masks to suppress the background noise, which is supervised by the recognition loss apart from the detection loss. 
As shown in the upper right corner of Fig.~\ref{fig:recognition conversion}, $M_1$ suppresses more background noise than $Mask_{K}$, where $M_1$ has higher activation in the text region and is lower in the background.
Therefore the masks $\{M_3, M_2, M_1\}$ produced by RC, which will be applied to the recognition features $\{r_3,r_2,r_1\}$, make the recognizer easier to concentrate on the text regions. 
% }

With RC, the gradient of recognition loss not only flows back to the backbone network but also to the proposal features. 
Optimized by both detection supervision and recognition supervision, the proposal features can better encode the high-level semantic information of the texts.
Therefore, the proposed RC can incentivize the coordination between detection and recognition.  

\subsection{Recognizer}
\label{sec:recognizer}
The RA dynamically samples text features, ensuring comprehensive coverage of all characters within the text area. Meanwhile, the RC suppresses the background noise in the recognition features. By incorporating RA and RC, our approach can achieve promising recognition performance through a sequential recognition network, eliminating the need for rectification techniques such as TPS ~\citep{bookstein1989principal}, RoISlide ~\citep{feng2019textdragon}, Bezier-Align ~\citep{liu2020abcnet} or character-level segmentation branch used in MaskTextSpotter~\citep{liao2019mask}. Inspired by ~\citep{yang2021focal}, we adopt a two-level self-attention mechanism (TLSAM) as the recognition encoder. Specifically, TLSAM consists of a fine-grained self-attention for capturing details from local areas and a coarse-grained self-attention for analyzing broader, more global patterns. During the decoding phase, a Spatial Attention Module (SAM)~\citep{liao2019mask} is utilized to output the recognition results. The Spatial Attention Module leverages positional encoding and a spatial attention mechanism with a Gated Recurrent Unit (GRU). When features are fed into the recognizer, positional encoding is first applied to these features. By adding positional information to each element in the input sequence, the attention mechanism can understand the relative positions of the elements. Then, the attention mechanism dynamically weights the importance of different parts of the input based on the current decoding state and outputs the recognition results. The overall recognition loss is defined as:
\begin{equation}
L_{reg} = -\frac{1}{T} \sum_{k=1}^{T}log\ p(y_i),
\end{equation}
wherein $T$ is the max length of the sequence and $p(y_i)$ is the probability of sequence.

\begin{table*}[t!]
\centering
\caption{End-to-end recognition result on RoIC13. P, R, and H represent precision,
recall, and Hmean, respectively. Bold denotes SOTA. Underline indicates second best. * represents adding TextOCR as part of the training set.}
\setlength{\tabcolsep}{5mm}{
\begin{tabular}{l|c|c|c|c|c|c}
\hline
\multirow{2}*{Method} & \multicolumn{3}{c|}{Rotation Angle $45^{\circ}$}                                            & \multicolumn{3}{c}{Rotation Angle $60^{\circ}$}  \\ \cline{2-7} 
                                                       & R                         & P                         & H                         & R           & P           & H          \\ \hline
CharNet~\citep{xing2019convolutional}                                                & 35.5                      & 34.2                      & 33.9                      & 8.4         & 10.3        & 9.3        \\ 
Mask TextSpotter~\citep{liao2019mask}                                       & \multicolumn{1}{c|}{45.8} & \multicolumn{1}{c|}{66.4} & \multicolumn{1}{c|}{54.2} & 48.3        & 68.2        & 56.6       \\ 
Mask TextSpotter v3~\citep{liao2020mask}                                    & 66.8                      & \textbf{88.5}                      & 76.1                      & 67.6        & \textbf{88.5}        & 76.6       \\
TTS~\citep{kittenplon2022towards}                                    & -                      & -                      & 80.4                      & -        & -        & 80.1       \\ 
LMTextSpotter~\citep{xia2024lmtextspotter}                                                & \textbf{86.1}  & 75.2 & 80.3                      & \textbf{85.6}  &  73.1 &  78.9      \\ \hline
SwinTextSpotter v2                                                  & 78.2                      & \underline{85.5}                      & \underline{81.7}                      & 79.3        & \underline{86.3}        & \underline{82.6}      \\
SwinTextSpotter v2*                                                  & \underline{80.8}                      & 84.7                      & \textbf{82.8}                      & \underline{81.0}        & 85.4        & \textbf{83.1} 

\\ \hline
\end{tabular}}
\label{Rotate-IC13-end-to-end}
\end{table*}

\begin{table}[t]
\centering
\caption{End-to-end recognition result on ICDAR 2015. “S”, “W”, and “G” represent recognition with “Strong”, “Weak”, and “Generic” lexicon, respectively. Bold denotes SOTA. Underline indicates second best. * represents adding TextOCR as part of the training set.}
\resizebox{\linewidth}{!}{%
\begin{tabular}{l|ccc}
\hline
\multirow{2}{*}{Method} & \multicolumn{3}{c}{ICDAR 2015 End-to-End}                    \\ \cline{2-4} 
                        & \multicolumn{1}{c|}{S}    & \multicolumn{1}{c|}{W}    & G    \\ \hline
FOTS~\citep{liu2018fots}                    & \multicolumn{1}{c|}{81.1} & \multicolumn{1}{c|}{75.9} & 60.8 \\
Mask TextSpotter~\citep{liao2019mask}        & \multicolumn{1}{c|}{83.0} & \multicolumn{1}{c|}{77.7} & 73.5 \\
CharNet~\citep{xing2019convolutional}                 & \multicolumn{1}{c|}{83.1} & \multicolumn{1}{c|}{79.2} & 69.1 \\
TextDragon~\citep{feng2019textdragon}              & \multicolumn{1}{c|}{82.5} & \multicolumn{1}{c|}{78.3} & 65.2 \\
Mask TextSpotter v3~\citep{liao2020mask}     & \multicolumn{1}{c|}{83.3} & \multicolumn{1}{c|}{78.1} & 74.2 \\
MANGO~\citep{qiao2021mango}                   & \multicolumn{1}{c|}{81.8} & \multicolumn{1}{c|}{78.9} & 67.3 \\
Mask TTD~\citep{feng2021residual}  & \multicolumn{1}{c|}{83.7} & \multicolumn{1}{c|}{79.2} & 65.8 \\
PAN++~\citep{wang2021pan++}                   & \multicolumn{1}{c|}{82.7} & \multicolumn{1}{c|}{78.2} & 69.2 \\
ABCNet v2~\citep{liu2021abcnetv2}                & \multicolumn{1}{c|}{82.7} & \multicolumn{1}{c|}{78.5} & 73.0 \\ 
TESTR~\citep{zhang2022text}               & \multicolumn{1}{c|}{85.2} & \multicolumn{1}{c|}{79.4} & 73.6 \\ 
TTS~\citep{kittenplon2022towards}               & \multicolumn{1}{c|}{85.2} & \multicolumn{1}{c|}{81.7} & \underline{77.4} \\ 
GLASS~\citep{ronen2022glass}  & \multicolumn{1}{c|}{84.7} & \multicolumn{1}{c|}{80.1} & 76.3 \\ 
SPTS~\citep{peng2022spts}               & \multicolumn{1}{c|}{77.5} & \multicolumn{1}{c|}{70.2} & 65.8 \\ 
ABINet++~\citep{fangabinet++}  & \multicolumn{1}{c|}{84.1} & \multicolumn{1}{c|}{80.4}  & \multicolumn{1}{c}{75.4} \\
SPTS v2~\citep{liu2023spts}               & \multicolumn{1}{c|}{81.2} & \multicolumn{1}{c|}{74.3} & 68.0 \\ 
DeepSolo~\citep{ye2023deepsolo}               & \multicolumn{1}{c|}{86.8} & \multicolumn{1}{c|}{81.9} & 76.9 \\ 
FastTCM~\citep{yu2024turning}               & \multicolumn{1}{c|}{\underline{87.0}} & \multicolumn{1}{c|}{\underline{82.0}} & 77.3 \\ 
FastTextSpotter~\citep{das2025fasttextspotter}              & \multicolumn{1}{c|}{86.6} & \multicolumn{1}{c|}{81.7} & 75.4 \\ 
IAST~\citep{zhang2024inverse}               & \multicolumn{1}{c|}{84.4} & \multicolumn{1}{c|}{80.0} & 73.8 \\ \hline
SwinTextSpotter         & \multicolumn{1}{c|}{83.9} & \multicolumn{1}{c|}{77.3} & 70.5 \\
SwinTextSpotter v2         & \multicolumn{1}{c|}{86.0} & \multicolumn{1}{c|}{80.7} & 75.4 \\ 
SwinTextSpotter v2*         & \multicolumn{1}{c|}{\textbf{89.6}} & \multicolumn{1}{c|}{\textbf{84.1}} & \textbf{79.4} \\ \hline
\end{tabular}}
\label{ICDAR 2015 End-to-End recognition result}
\end{table}

\begin{table}[t]
\centering
\caption{End-to-end text spotting result and detection result on ReCTS. Bold denotes SOTA. Underline indicates second best.}
\resizebox{\linewidth}{!}{%
\begin{tabular}{l|c|c|c|c}
\hline
\multirow{2}{*}{Method} & \multicolumn{3}{c|}{Detection} & \multirow{2}{*}{1-NED} \\ \cline{2-4}
                        & R        & P        & H        &                        \\ \hline
FOTS~\citep{liu2018fots}                    & 82.5     & 78.3     & 80.31    & 50.8                   \\ 
MaskTextSpotter~\citep{liao2019mask}         & 88.8     & 89.3     & 89.0       & 67.8                   \\ 
AE TextSpotter~\citep{wang2020ae}          & \underline{91.0}     & 92.6     & \underline{91.8}     & 71.8                   \\ 
ABCNet v2~\citep{liu2021abcnetv2}               & 87.5     & \underline{93.6}     & 90.4     & 62.7                   \\ 
ABINet++~\citep{fangabinet++}               & 89.2     & 92.7     & 90.9     & \textbf{76.5}                   \\  \hline
SwinTextSpotter                    & 87.1     & \textbf{94.1}     & 90.4     & \underline{72.5}                   \\ 
SwinTextSpotter v2                    & \textbf{91.1}     & 93.0     & \textbf{92.1}     & \textbf{76.5}                   \\ \hline
\end{tabular}}
\label{ReCTS Rsult}
\end{table}

\begin{table}[t]
\centering
\caption{End-to-end text spotting result on VinText. ABCNet+D means adding the methods proposed in ~\citep{m_Nguyen-etal-CVPR21} to ABCNet. The same to Mask Textspotter v3+D. Bold denotes SOTA. Underline indicates second best. * represents adding TextOCR as part of the training set.}
\begin{tabular}{l|c}
\hline
Method & H-mean                       \\ \hline 
ABCNet~\citep{liu2020abcnet}                  & 54.2                     \\
ABCNet+D~\citep{m_Nguyen-etal-CVPR21}     & 57.4                     \\ 
Mask Textspotter v3~\citep{m_Nguyen-etal-CVPR21}     & 53.4                     \\ 
Mask Textspotter v3+D~\citep{m_Nguyen-etal-CVPR21}     & 68.5                     \\ 
FastTextSpotter~\citep{das2025fasttextspotter} & 73.0                     \\ 
SwinTextSpotter                    & 71.1                        \\ 
SwinTextSpotter v2                    & \underline{73.1}                        \\ 
SwinTextSpotter v2*                   & \textbf{74.9}                        \\\hline 
\end{tabular}
\label{VinText}
\end{table}

\begin{table}[t]
\centering
\caption{End-to-end text spotting result and detection result on SCUT-CTW1500.  ``None"
represents lexicon-free. ``Full" represents that we use all the words that appeared in the test set. Bold denotes SOTA. Underline indicates second best. * represents adding TextOCR as part of the training set.}
\begin{tabular}{l|c|c|c}
\hline
\multirow{2}*{Method} & Detection & \multicolumn{2}{c}{End-to-End}                      \\ \cline{2-4} 
            & H-mean           & None                      & \multicolumn{1}{c}{Full} \\ \hline
TextDragon~\citep{feng2019textdragon}       & 83.6           & 39.7                      & 72.4                   \\ 
ABCNet~\citep{liu2020abcnet}     & 81.4            & 45.2                      & \multicolumn{1}{c}{74.1}  \\
MANGO~\citep{qiao2021mango}      & -            & 58.9                      & 78.7                   \\ 
Mask TTD~\citep{feng2021residual} & 85.0     & 42.2   & 74.9                \\ 
ABCNet v2~\citep{liu2021abcnetv2}   & 84.7            & 57.5                      & 77.2                   \\
TESTR~\citep{zhang2022text}   & 87.1       &56.0  & 81.5
                   \\
SPTS~\citep{peng2022spts}   & -            & \underline{63.6} & \underline{83.8}                   \\
ABINet++~\citep{fangabinet++}   & -            &  60.2 & 80.3                   \\ 
SPTS v2~\citep{liu2023spts}   & -            & \textbf{64.4} & \textbf{84.0}       \\
DeepSolo~\citep{ye2023deepsolo}   & -  & 60.1 & 78.4 \\
FastTCM~\citep{yu2024turning} & -  & 60.4 & 78.8 \\

\hline
SwinTextSpotter         & \underline{88.0}           & 51.8                      & \multicolumn{1}{c}{77.0}   \\ 
SwinTextSpotter v2       & 88.2           & 57.0                    & \multicolumn{1}{c}{77.5} \\
SwinTextSpotter v2*         & \textbf{89.3}           & 61.3                    & \multicolumn{1}{c}{82.0}
\\ \hline
\end{tabular}
\label{tab:ctw1500_e2e}
\end{table}

\section{Experiments}
We try to investigate the effectiveness of our method on seven scene text benchmarks, including two multi-oriented scene text benchmarks ICDAR 2015~\citep{karatzas2015icdar} and RoIC13~\citep{liao2020mask}, two multilingual benchmarks ReCTS~\citep{zhang2019icdar} and Vintext~\citep{m_Nguyen-etal-CVPR21}, and three arbitrarily-shaped scene text benchmarks Total-Text~\citep{ch2020total}, SCUT-CTW 1500~\citep{liu2019curved} and Inverse-Text~\citep{ye2022dptext}.To evaluate the effectiveness of each component of our proposed method, we conduct ablation studies on Total-Text. Unless specified, all values in the tables are in percentage.

\subsection{Datasets}

We use the following datasets to train the model.

\textbf{Curved SynthText}~\citep{liu2020abcnet} is a synthesized dataset specifically designed for arbitrarily-shaped scene text, compared to the previous SynthText~\citep{gupta2016synthetic} which only focused on regular texts. It consists of 94,723 images with multi-oriented text and 54,327 images with curved text.

\textbf{ICDAR 2013}~\citep{karatzas2013icdar} is a scene text dataset proposed in 2013. It contains 229 training images and 233 test images.

\textbf{ICDAR 2015~\citep{karatzas2015icdar}} is a scene text dataset that was created in 2015. It comprises a total of 1,500 images, with 1,000 images designated for training and 500 images for testing. These images contain scene text with various orientations, font styles, and scales.

\textbf{ICDAR 2017}~\citep{nayef2017icdar2017} is a multi-lingual dataset that includes 7,200 training images and 1,800 validation images. For the purpose of this study, only images containing English text were selected for training.

\textbf{ICDAR19 ArT}~\citep{chng2019icdar2019} is a dataset for arbitrarily shaped text. It contains 5,603 training images.

\textbf{ICDAR19 LSVT}~\citep{sun2019icdar} is a large number of Chinese datasets which contain 30,000 training images.

\textbf{Total-Text}~\citep{ch2020total} is a widely used benchmark for the evaluation of arbitrarily-shaped scene text detection and text spotting. It consists of 1,255 training images and 300 testing images. The word-level polygon boxes are provided as annotations.

\textbf{SCUT-CTW1500}~\citep{liu2019curved} is a text-line level arbitrarily-shaped scene text dataset. It consists of 1,000 training images and 500 testing images. Compared to Total-Text, this dataset contains the denser and longer text.

\textbf{ReCTS}~\citep{zhang2019icdar} is a large-scale Chinese scene text dataset that contains 20,000 training images and 5,000 testing images. It further provides character-level annotations, but these annotations are not used in our proposed method.

\textbf{VinText}~\citep{m_Nguyen-etal-CVPR21} is a recently proposed Vietnamese text dataset. It consists of 1,200 training images and 500 testing images.

\textbf{TextOCR}~\citep{singh2021textocr} is a large-scale English scene text dataset which contains 21,778 training and 3,124 validation images.

\textbf{Inverse-Text}~\citep{ye2022dptext} is an arbitrary-shape scene text with about 40\% inverse-like instances. It consists of 500 testing images without train data.

\subsection{Metrics}
\label{metrics}
\textbf{Detection.} 
Following previous works~\citep{feng2019textdragon,liao2020mask,liu2021abcnetv2,zhang2022text}, the Hmean (F-measure) is used as the metric for evaluating the detection performance~\citep{wolf2006object,karatzas2015icdar,ch2020total,liu2019curved,m_Nguyen-etal-CVPR21}. It is calculated based on both the recall and precision of the detected word polygons in comparison to the ground truth polygons. A detection result is considered as the true positive (TP) if the detection result has more than 0.5 IoU (intersection over union) with the ground truth Polygon. Precision, Recall, and H-mean are calculated as follows:
\begin{equation}
Precision=\frac{TP}{TP+FP} ,\\
\end{equation}
\begin{equation}
Recall=\frac{TP}{TP+FN} ,\\
\end{equation}
\begin{equation}
Hmean=\frac{2*Precision*Recall}{Precision+Recall} ,
\end{equation}
where TP, FP, and FN denote true positive, false positive, and false negative, respectively.

\textbf{Text Spotting.} Following previous works~\citep{feng2019textdragon,liao2020mask,liu2021abcnetv2,zhang2022text}, for English benchmarks and Vietnamese benchmarks, Hmean is used as the metric for evaluating the text spotting performance. The calculation method is similar to the detection part, with the difference being that only those with an IoU greater than 0.5 and correct recognition results are considered true positive (TP). The evaluation protocol for the Chinese benchmark involves a two-step process~\citep{zhang2019icdar}. Initially, each detected text region is paired with the ground-truth polygon that has the maximum IOU, or it is matched to `None’ if none IOU is larger than 0.5. Subsequently, the normalized edit distance is computed using all successfully matched pairs ($t_i \hat{t}_i$), according to the following formula: 
\begin{equation}
\mbox{accuracy} = 1 - \frac{1}{N} \sum_{i=1}^{N} \frac{D(t_i, \hat{t}_i)}{\max(t_i, \hat{t}_i)}, \\
\end{equation}
where D represents the Levenshtein Distance, $t_i$ represents the predicted text. $\hat{t}_i$ represents the corresponding ground truth. N is the total number of text.

\subsection{Implementation Details}
\label{Implementation Details}

We follow the training strategy in ~\citep{li2017towards,liu2018fots,liao2019mask,liao2020mask}. First, the model is pretrained on the Curved SynthText ~\citep{liu2020abcnet}, ICDAR-MLT~\citep{nayef2017icdar2017}, and the corresponding dataset for 450k iterations. The initialized learning rate is $2.5 \times 10 ^ {-5}$, which reduces to $2.5 \times 10 ^ {-6}$ at $380k ^ {th}$ iteration and $2.5 \times 10 ^ {-7}$ at $420k ^ {th}$ iteration. Then we jointly train the pretrained model for 120k iterations on the Total-Text, ICDAR 2015, and ICDAR-MLT, which decays to a tenth at 90k. For Chinese, We follow the training strategies in ~\citep{liu2021abcnetv2}. We adopt the Chinese synthetic pretrained data~~\citep{liu2021abcnetv2}, ICDAR19 ArT, ICDAR19 LSVT and ReCTS to pretrain the model. Then we finetune the pretrained model on the ReCTS. The experimental conditions on Inverse-Text are similar to the previous method~\citep{ye2022dptext}. We initiate the process by training our model on the dataset detailed in Table~\ref{tab:inverse_e2e}. Subsequently, we conduct fine-tuning using the Total-Text dataset to evaluate the model on Inverse-Text. The metrics to evaluate Inverse-Text are presented in Section~\ref{metrics}. For Vintext, we follow the training strategies~\citep{m_Nguyen-etal-CVPR21,huang2022swintextspotter} to train the model. Following previous methods~\citep{carion2020end,zhu2020deformable,sun2021sparse}, we set the value of K in the BSS to 300.

We extract 4 feature maps with 1/4, 1/8, 1/16, and 1/32 resolution of the input image for text detection and 1/4 resolution of the input image for text recognition. We train our model with an image batch size of 8. The following data augmentation strategies are used:  
(1) random scaling with the short size chosen from 640 to 896 (interval of 32) and the long size being less than 1600; (2) random rotation which rotates the images in an angle range of [$-90 ^ {\circ}$, $90 ^ {\circ}$]; and (3) random crop which is randoms crop the images with text. Other strategies such as random brightness, contrast, and saturation are also applied during training.

\begin{table*}[!th]
\centering
\caption{\small End-to-end text spotting result and detection result on Total-Text.  ``None" represents lexicon-free. ``Full" represents that we use all the words appeared in the test set. Bold denotes SOTA. Underline indicates second best.}
\vspace{-0.3em}
\setlength{\tabcolsep}{10pt}
\resizebox{\linewidth}{!}{%
\begin{tabular}{l|c|c|c|c|c|c}
\hline
\multirow{2}*{Method} & \multirow{2}*{External Data} & \multicolumn{3}{c|}{Detection} & \multicolumn{2}{c}{End-to-End}                      \\ \cline{3-7}
    & & P  & R    & H             & None                      & \multicolumn{1}{c}{Full} \\ \hline
Mask TextSpotter~\citep{lyu2018mask} & Synth800K+IC13+IC15+SCUT  & 69.0 & 55.0 &  61.3 &  52.9 & 71.8 \\
CharNet~\citep{xing2019convolutional} & Synth800K & 87.3 & 85.0  & 86.1 & 66.2 & $-$ \\
Text Dragon~\citep{feng2019textdragon} & Synth800K  & 85.6 & 75.7 & 80.3 & 48.8 & 74.8 \\
Boundary TextSpotter~\citep{wang2020all} & Synth800K+IC13+IC15  & 88.9 & 85.0 & 87.0 & 65.0 & 76.1  \\
Text Perceptron~\citep{qiao2020text} & Synth800K & 88.8 & 81.8 & 85.2 & 69.7 & 78.3 \\
CRAFTS~\citep{baek2020character} & Synth800K+IC13  & 89.5 & 85.4 & 87.4 & 78.7 & - \\
Mask TextSpotter v3~\citep{liao2020mask} &Synth800K+IC13+IC15+SCUT   & $-$ & $-$ & $-$ & 71.2 & 78.4  \\ 
Mask TTD~\citep{feng2021residual} & Synth800K & 87.1 & 80.3 & 83.5 & 54.6 & 78.6  \\ 
ABCNet v2~\citep{liu2021abcnetv2} & Synth150K+MLT17 & 90.2 & 84.1  & 87.0 & 70.4 & 78.1  \\ 
MANGO~\citep{qiao2021mango} & Synth800K+Synth150K+COCO-Text+MLT19+IC13+Total-Text & $-$ & $-$ & $-$ & 72.9 & 83.6  \\
PGNet~\citep{wang2021pgnet} & Synth800K+IC15 & 85.5 & \underline{86.8}   & 86.1 & 63.1 & $-$ \\
TESTR~\citep{zhang2022text} & Synth150K+MLT17 & \textbf{93.4} & 81.4 & 86.9 & 73.3 & 83.9  \\
TTS~\citep{kittenplon2022towards} & Synth800K+COCO-Text+IC13+IC15+SCUT & $-$ & $-$ & $-$ & 78.2 & 86.3  \\
GLASS~\citep{ronen2022glass} & Synth800K & $-$ & $-$ & $-$ & \underline{79.9} & 86.2  \\
SPTS~\citep{peng2022spts} & Synth150K+MLT17+IC13+IC15 & $-$ & $-$ & $-$ & 74.2 & 82.4  \\
ABINet++~\citep{fangabinet++} & Synth150K+MLT17+IC15 & $-$ & $-$ & $-$ & 77.6 & 84.5  \\ 
SPTS v2~\citep{liu2023spts} & Synth150K+MLT17+IC13+IC15 & $-$ & $-$ & $-$ & 75.0 & 82.6  \\ 
DeepSolo~\citep{ye2023deepsolo} & Synth150K+MLT17+IC13+IC15 & 93.1 & 82.1 & 87.3 & 79.7 & 87.0  \\
UNITS~\citep{kil2023towards} & Synth150K+MLT17+IC13+IC15+Total-Text+TextOCR+HierText & - & - &89.8 & 78.7 & 86.0 \\
FastTCM~\citep{yu2024turning} & Synth150K+MLT17+IC13+IC15 &  -& - & 87.8 & \underline{79.9} & \underline{87.2}  \\
FastTextSpotter~\citep{das2025fasttextspotter} & Synth150K+MLT17 &  90.6 & 85.5 & 88.0 & 75.1 & 86.0 \\
IAST~\citep{zhang2024inverse} & Synth150K+MLT17+IC15 & 94.7 & 85.2 & \underline{89.7} & 71.9 & 83.5 \\
\hline
SwinTextSpotter-Res~\citep{huang2022swintextspotter} & Synth150K+MLT17+IC13+IC15   & $-$ & $-$  & 87.2             & 72.4                      & \multicolumn{1}{c}{83.0}   \\
SwinTextSpotter~\citep{huang2022swintextspotter} & Synth150K+MLT17+IC13+IC15  & $-$ & $-$ & 88.0 & 74.3 & 84.1  \\
SwinTextSpotter v2 & Synth150K+MLT17+IC13+IC15 & 91.1 & 84.5 & 87.7 & 78.6 & 86.3  \\
SwinTextSpotter v2 & Synth150K+MLT17+IC13+IC15+TextOCR & \underline{93.1} & \textbf{87.2} & \textbf{90.1} & \textbf{82.8} & \textbf{88.4} \\
\hline
\label{tab:totaltext_e2e}
\end{tabular}
}
\end{table*}

\begin{table*}[!th]
\centering
\caption{\small End-to-end text spotting result and detection result on Inverse-Text.  ``None" represents lexicon-free. ``Full" represents that we use all the words that appeared in the test set. Bold denotes SOTA. Underline indicates second best.}
\vspace{-0.3em}
\setlength{\tabcolsep}{10pt}
\resizebox{\linewidth}{!}{%
\begin{tabular}{l|c|c|c|c|c|c}
\hline
\multirow{2}*{Method} & \multirow{2}*{Training Data} & \multicolumn{3}{c|}{Detection} & \multicolumn{2}{c}{End-to-End}                      \\ \cline{3-7}
    &  & P  & R    & H             & None                      & \multicolumn{1}{c}{Full} \\ \hline
Mask TextSpotter v2~\citep{liao2019mask} & Synth800K+IC13+IC15+SCUT+Total-Text & - & - & - & 39.0 & 43.5
  \\ 
ABCNet~\citep{liu2020abcnet} & Synth150K+MLT17+Total-Text & - & - & - & 22.2 & 34.3
  \\ 
ABCNet v2~\citep{liu2021abcnetv2} & Synth150K+MLT17+Total-Text & 82.0 & 70.2 & 75.6 & 34.5 & 47.4
  \\ 
TESTR~\citep{zhang2022text} & Synth150K+MLT17+IC15+Total-Text & 83.1 & 67.4 & 74.4 & 34.2 & 41.6  \\
SPTS~\citep{peng2022spts} & Synth150K+MLT17+IC13+IC15+Total-Text & - & - & - & 38.3 & 46.2  \\
ABINet++~\citep{fangabinet++} & Synth150K+MLT17+IC15+Total-Text & 78.5 & 65.4 & 71.3 & 43.7 & 48.1  \\
DeepSolo~\citep{ye2023deepsolo} & Synth150K+MLT17+IC15+IC13+Total-Text & - & - & - & 48.5 & 53.9  \\
IAST~\citep{zhang2024inverse} & Synth150K+MLT17+IC15+Total-Text &  92.5 & 86.6 & 89.5 & \underline{68.8} & \underline{80.6}  \\
\hline
SwinTextSpotter~\citep{huang2022swintextspotter} & Synth150K+MLT17+IC13+IC15+Total-Text &  \textbf{94.5} & 85.8 & 89.9 & 55.4 & 67.9  \\
SwinTextSpotter v2 & Synth150K+MLT17+IC13+IC15+Total-Text & \underline{93.6} & \underline{87.1} & \underline{90.2} & 64.8 & 76.5 \\
SwinTextSpotter v2 & Synth150K+MLT17+IC13+IC15+Total-Text+TextOCR & 94.1 & \textbf{88.0} & \textbf{90.9} & \textbf{69.4} & \textbf{81.2} \\
\hline
\label{tab:inverse_e2e}
\end{tabular}
}
\end{table*}

\begin{figure*}[ht!]
    \centering
    \begin{minipage}[c]{0.45\linewidth}
        \includegraphics[width=8.0cm, height=4.3cm]{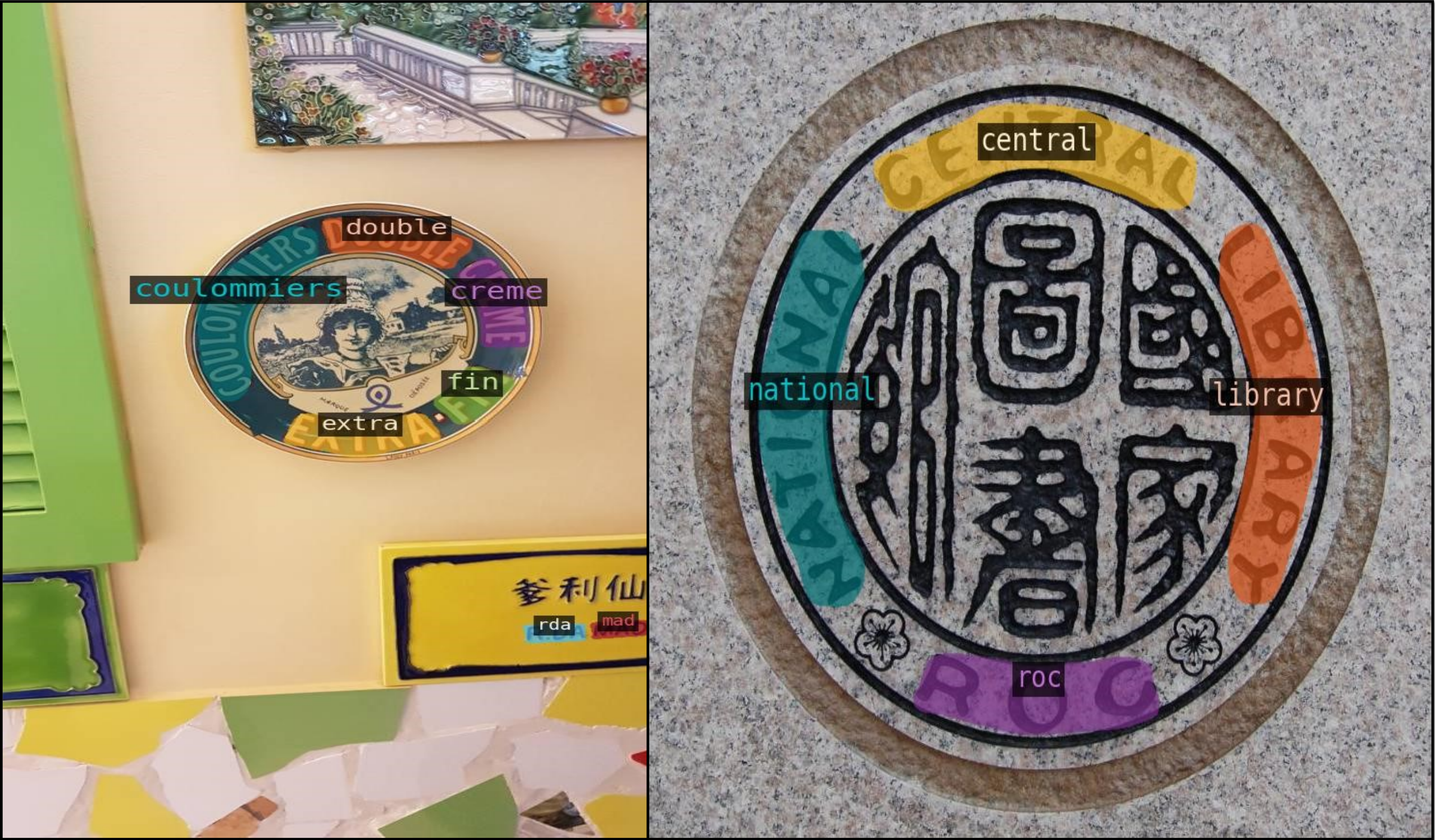}
        \centerline{(a) Total-Text}
    \end{minipage}
    \begin{minipage}[c]{0.45\linewidth}
        \includegraphics[width=8.0cm, height=4.3cm]{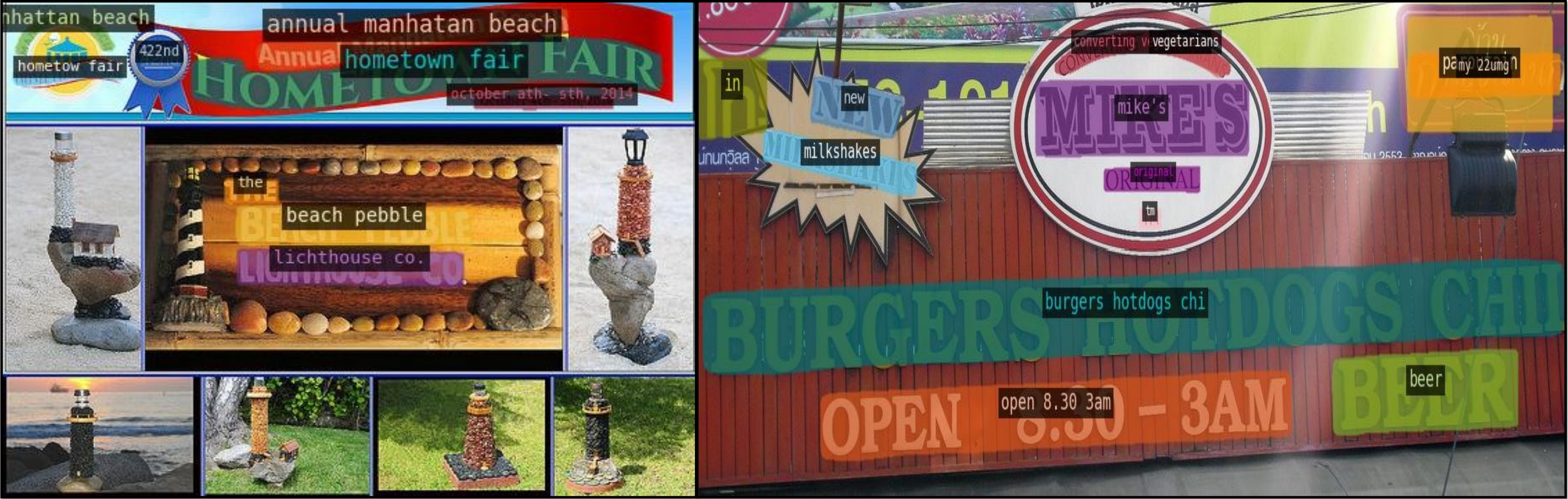}
        \centerline{(b) CTW1500}
    \end{minipage}
    \begin{minipage}[c]{0.45\linewidth}
        \includegraphics[width=8.0cm, height=4.3cm]{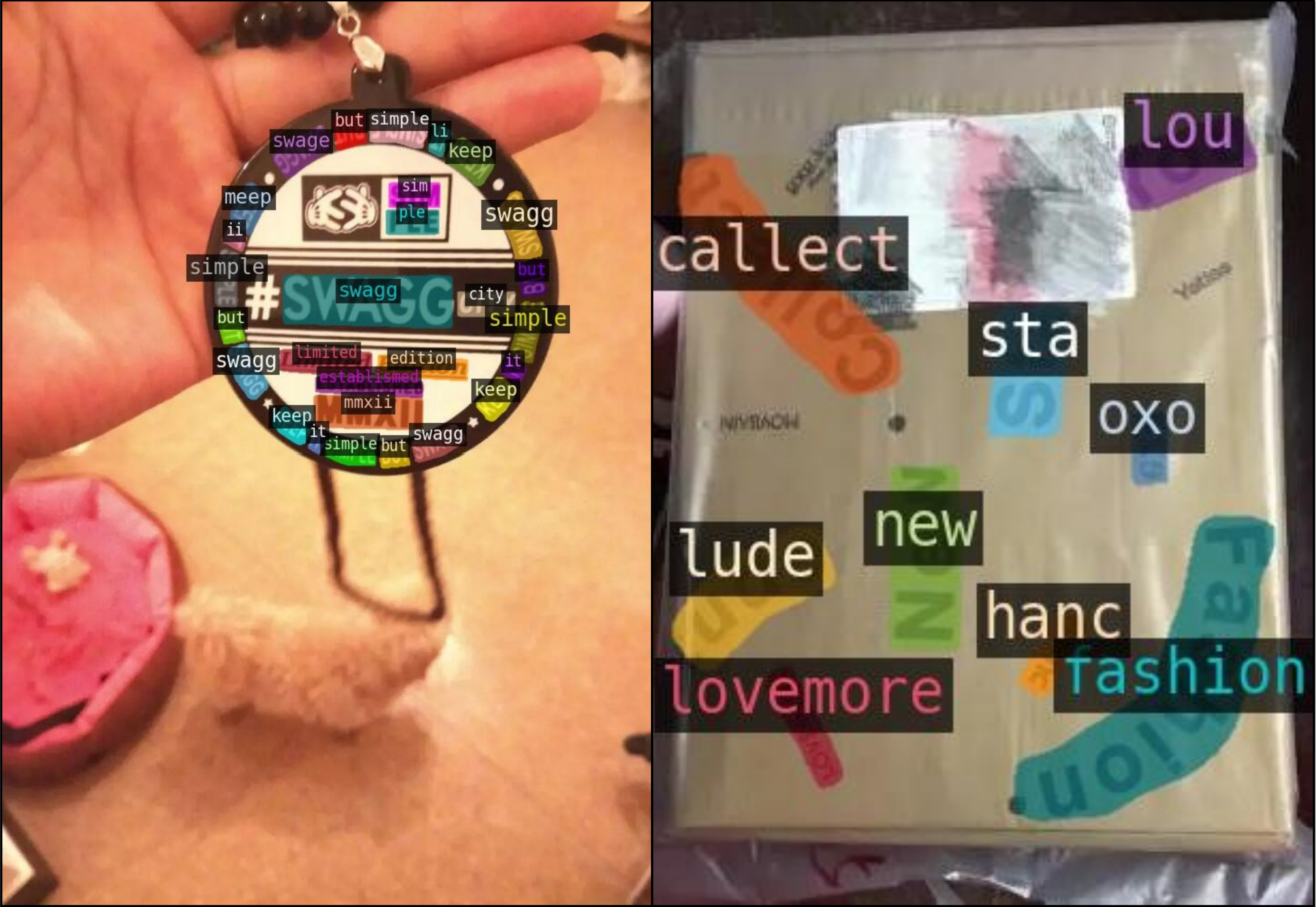}
        \centerline{(e) Inverse-Text}
    \end{minipage}
    \begin{minipage}[c]{0.45\linewidth}
        \includegraphics[width=8.0cm, height=4.3cm]{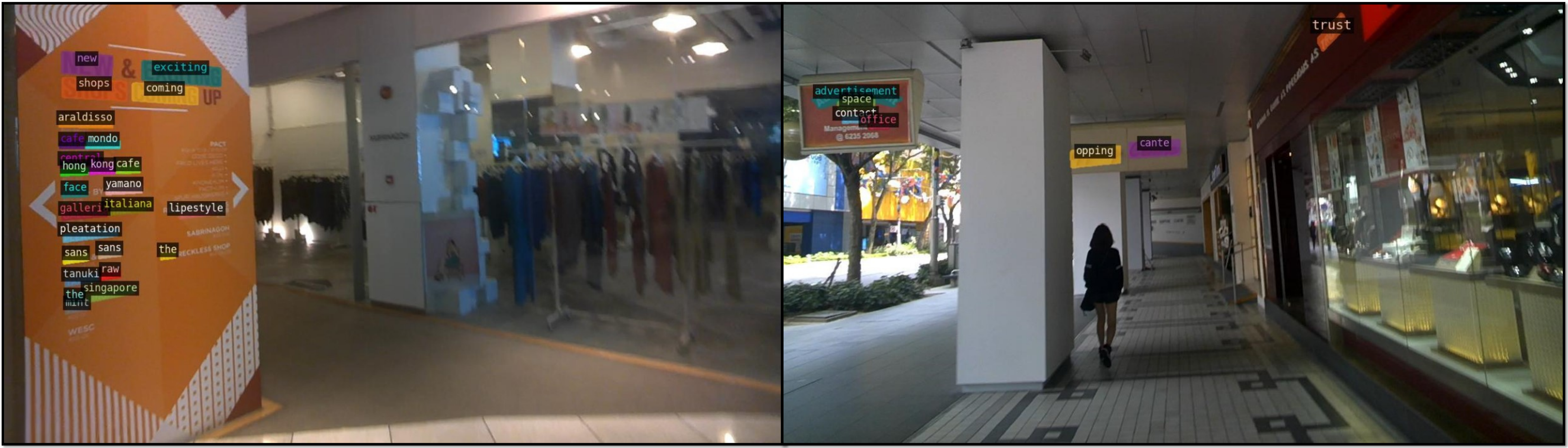}
        \centerline{(c) ICDAR2015}
    \end{minipage}
    \begin{minipage}[c]{0.45\linewidth}
        \includegraphics[width=8.0cm, height=4.3cm]{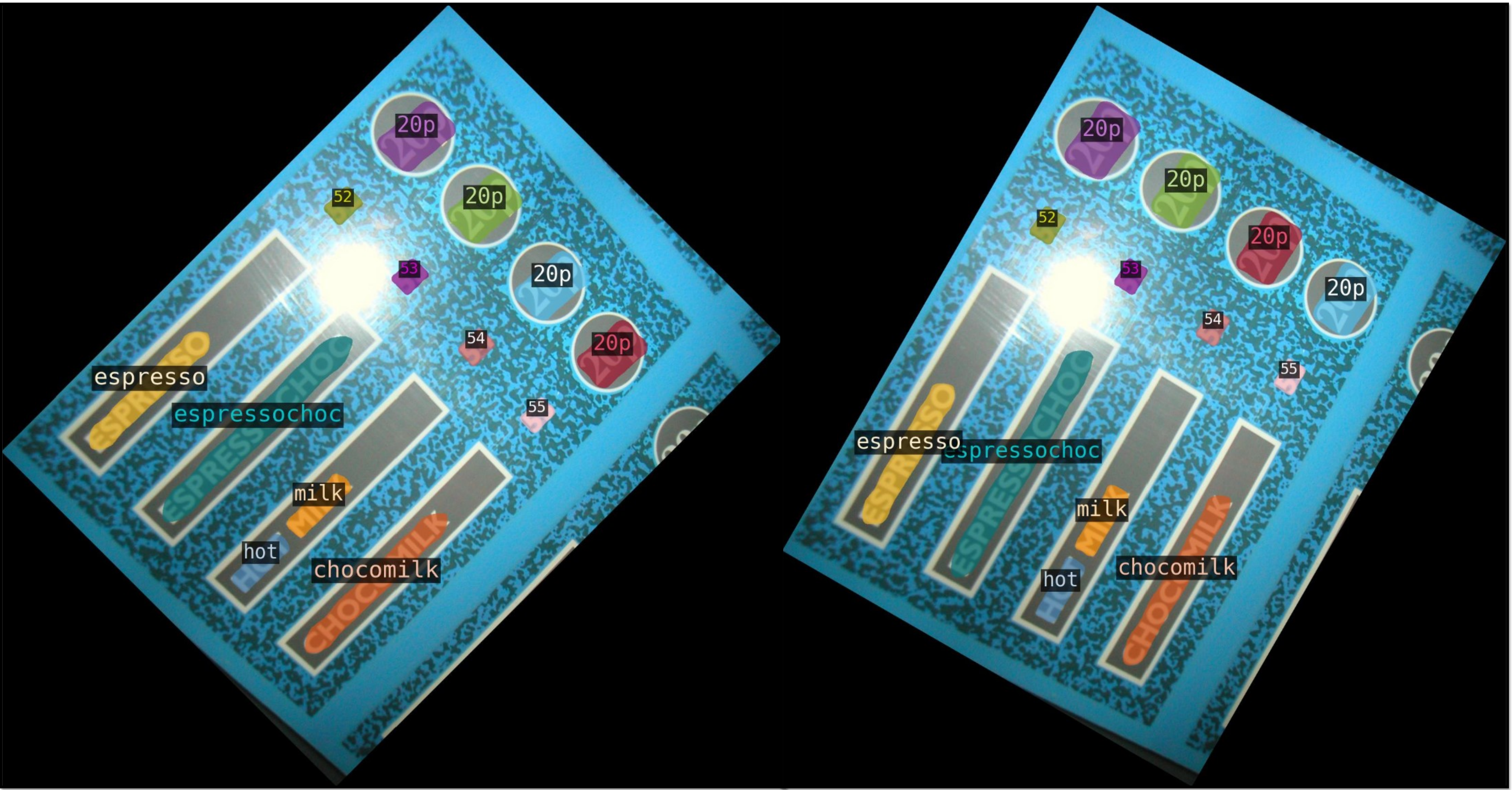}
        \centerline{(d) RoIC13}
    \end{minipage}
    \begin{minipage}[c]{0.45\linewidth}
        \includegraphics[width=8.0cm, height=4.3cm]{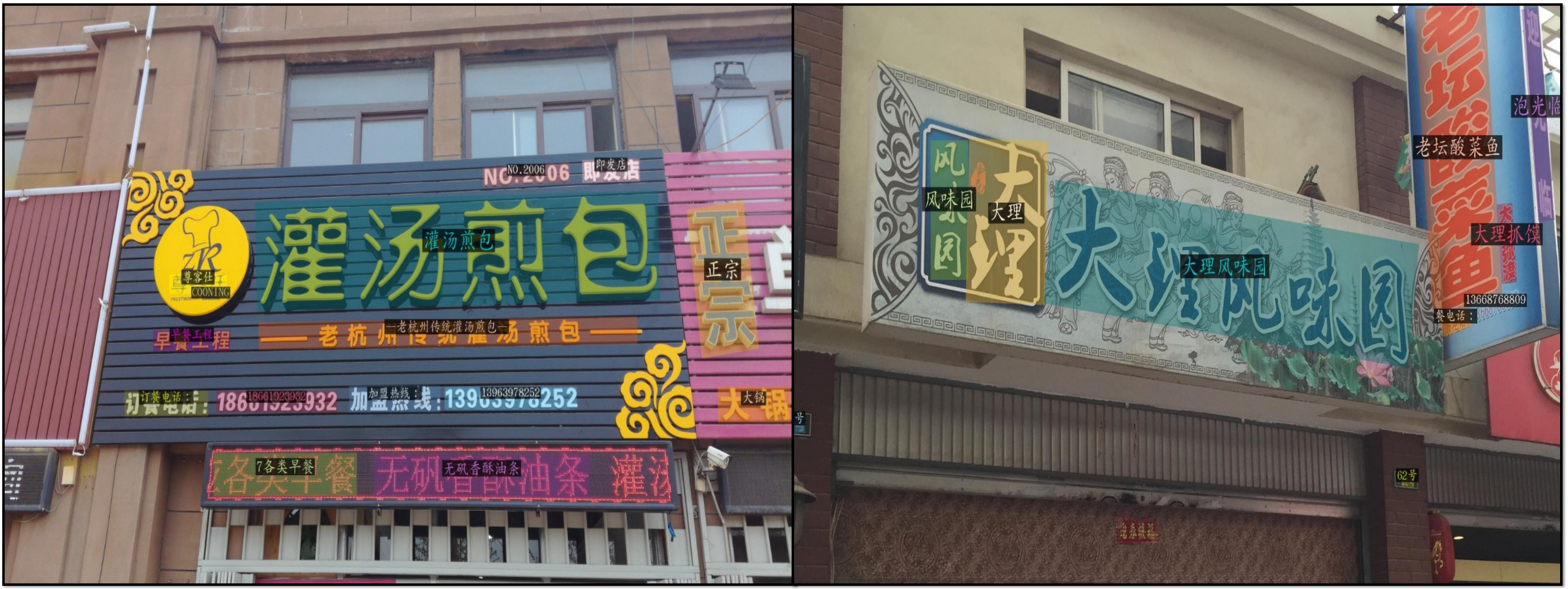}
        \centerline{(f) ReCTS}
    \end{minipage}
    \begin{minipage}[c]{0.45\linewidth}
        \includegraphics[width=8.0cm, height=4.3cm]{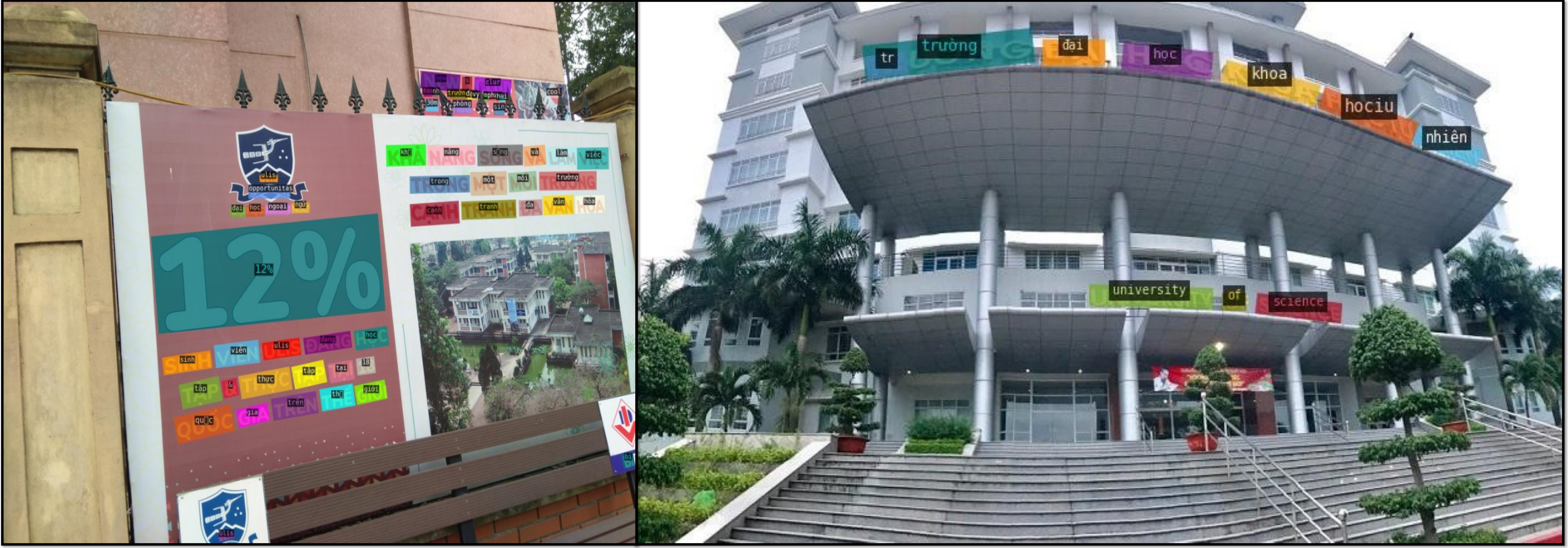}
        \centerline{(g) VinText}
    \end{minipage}
    \begin{minipage}[c]{0.45\linewidth}
        \includegraphics[width=8.0cm, height=4.3cm]{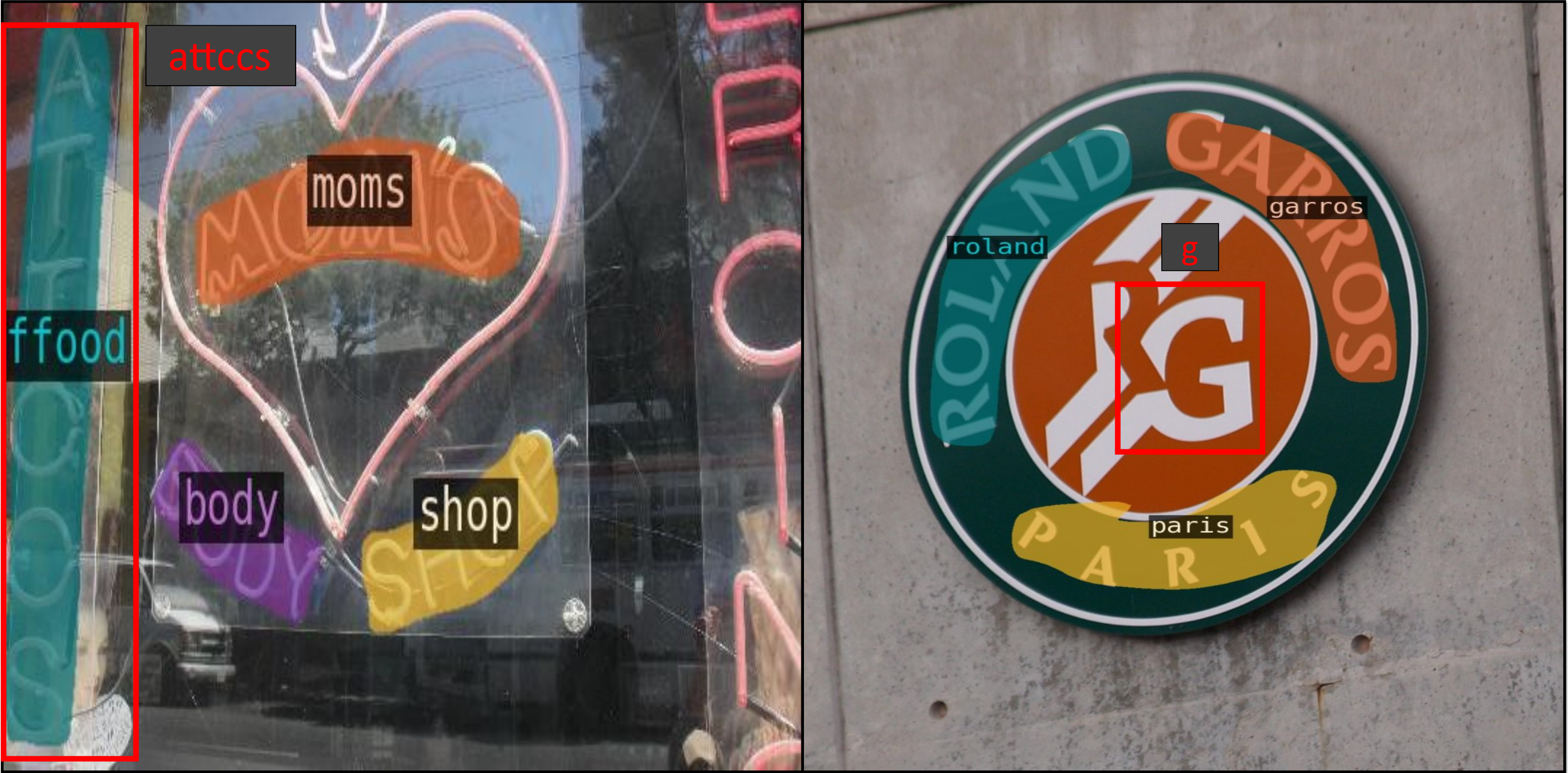}
        \centerline{(h) Failure examples of SwinTextSpotter v2}
    \end{minipage}
    \caption{Qualitative results on the scene text benchmarks. Images are selected from Total-Text (first row), SCUT-CTW1500 (second row), ICDAR 2013 (third row), ICDAR 2015 (fourth row), and Inverse-Text (fifth row). %Zoom in for best view.
    Best viewed on screen.
    }\label{fig:vis_results}
\end{figure*}

\subsection{Comparisons with State-of-the-Art methods}
Except in special cases, all values in the table are in percentage.

\textbf{Multi-oriented and Multilingual datasets}. To verify the effectiveness of SwinTextSpotter v2 on oriented scene text, we first conduct experiments on ICDAR 2015. The results are shown in Table~\ref{ICDAR 2015 End-to-End recognition result}. It can be observed that SwinTextSpotter v2 achieves state-of-the-art performance using both the ``Strong" and ``Weak" lexicons, without the use of character-level annotations and the weighted edit distance~\citep{liao2019mask} that were used in previous methods. Notably, SwinTextSpotter v2 also outperforms ABINet++~\citep{fangabinet++}, which employs an additional language model in the recognizer. To evaluate the rotation robustness of SwinTextSpotter v2, we test it on the RoIC13 dataset proposed in~\citep{liao2020mask}. The end-to-end recognition results are presented in Table~\ref{Rotate-IC13-end-to-end}. In both the Rotation Angle $45^{\circ}$ and Rotation Angle $60^{\circ}$ datasets, SwinTextSpotter v2 achieves state-of-the-art results in terms of the H-mean metric. TTS adopts about 866.6k images to train the model while our method utilizes 161.5k images to train the model. Our approach outperforms TTS by 1.3$\%$ in terms of H-mean on Rotation Angle $45^{\circ}$ and 2.5$\%$ on Rotation Angle $60^{\circ}$. Compared to the LMTextSpotter, our method outperforms it by 1.4$\%$ in terms of H-mean on Rotation Angle $45^{\circ}$ and 3.7$\%$ on Rotation Angle $60^{\circ}$. With the RA, the model can dynamically extract text features for recognition. Consequently, this leads to a more accurate recognition of text at various rotation angles.

To demonstrate the generalization ability of our method, we conduct further experiments on multilingual benchmarks including ReCTS and VinText. The results are presented in Table~\ref{ReCTS Rsult} for ReCTS. Our method outperforms ABCNet v2, which only works on word-level annotation, by 13.8$\%$ in terms of 1-NED. SwinTextSpotter v2 is also comparable to ABINet++ which adopts a language model to improve the recognition performance. Moreover, SwinTextSpotter v2 achieves 4.7$\%$ higher 1-NED than AE-TextSpotter, the state-of-the-art method that requires additional character-level annotation. For VinText, the end-to-end results are shown in Table~\ref{VinText}, where `D' denotes using a dictionary for the training of the recognizer. SwinTextSpotter v2 outperforms the SOTA method FastTextSpotter by $1.9\%$ on VinText, which further demonstrates the generalization ability of our method.
It is worth noting that for the above tasks, we do not use the dictionary for the training of recognizer as ABCNet+D and Mask TextSpotter v3+D.
The qualitative results of these three benchmarks are shown in Fig.~\ref{fig:vis_results} (d)(e)(f)(g).

\textbf{Irregular text}. We conduct experiments on three arbitrarily-shaped scene text datasets (Total-Text, SCUT-CTW1500, and Inverse-Text) for both detection and end-to-end scene text spotting tasks. In the text detection task, the results shown in Table~\ref{tab:totaltext_e2e}, Table~\ref{tab:ctw1500_e2e}, and Table~\ref{tab:inverse_e2e} demonstrate that SwinTextSpotter v2 can achieve good performance on both datasets. When using TextOCR~\citep{singh2021textocr}, SwinTextSpotter v2 significantly outperforms previous methods on Total-Text, with an 82.8\% on `H-mean' metric, which is 2.9\% higher than FastTCM and 3.1\% higher than DeepSolo. Although UNITS adopts more training data, our method surpasses it by 4.1\%. Additionally, our method enhances the synergy between detection and recognition which can improve both the detection and recognition performance. Our method also achieves the best detection performance with 90.1\% on the `H-mean' metric. On SCUT-CTW1500, as presented in Table~\ref{tab:ctw1500_e2e}, SwinTextSpotter v2 can achieve 89.3$\%$ on the detection performance, higher than the SOTA method by 2.2\%. On the Inverse-Text, when compared to the IAST, which is specifically designed for inverse-like text, our approach still outperforms it by $0.6\%$. Additionally, our method surpasses IAST by 10.9\% on the TotalText dataset and 5.6\% on the `General' metric of ICDAR2015. Compared to the ABINet++, our method surpasses it by 21.1\% on the `None' metric without using a language model to enhance the recognition. Previous methods lack synergy between detection and recognition, which could easily be misled by background noise and are not effective in detecting and recognizing highly curved text. In contrast, our method enhances the synergy between detection and recognition via the RC and more accurately extracts text features via RA. This advancement enables text spotters to accurately detect and recognize text instances, even in instances where the text is inverted. Some qualitative results are shown in Fig.~\ref{fig:vis_results} (a)(b)(c).

\textbf{Compared to the conference version}. In SwinTextSpotter v2, a 50\% reduction in parameters compared to its conference version boosts computational efficiency in real-world applications. Specifically, in the conference version, the inference speed of the model and FLOPs of the detector are 2.0 FPS and 28.385G, respectively. In contrast, SwinTextSpotter v2, with its 50\% parameter reduction, achieves double the speed of the conference version and about halves the FLOPs (4.0 FPS and 16.387G FLOPs). Furthermore, while reducing computational requirements, SwinTextSpotter v2 outperforms the conference version by 4.3\% on the `None' metric of TotalText and by 4.9\% on the `Generic' metric of ICDAR2015.

% In SwinTextSpotter v2, the synergy between detection and recognition has been enhanced through RC, and RA can more accurately extract text features, thereby improving the detection and recognition of text in any shape. These results demonstrate the significance of the better synergy between text detection and recognition, which enables text spotters to accurately detect and recognize text instances, even in instances where the text is inverted. Some qualitative results are shown in Fig.~\ref{fig:vis_results} (a)(b)(c).

\begin{table*}[h]
\centering
\caption{Ablation studies on Total-Text and ICDAR2015 without finetuning. TRS means SwinTextSpotter~\citep{huang2022swintextspotter} with three refinement stages and gets rid of the dilated convolution. BSS means the Box Selection Schedule. RA means Recognition Alignment. The parameters contain parameters of the detector and recognizer.}
\resizebox{0.9\linewidth}{!}{
\begin{tabular}{@{}l|ccc|cc|ccc|c@{}}
\hline 
\multirow{2}{*}{Method} & \multicolumn{1}{l|}{\multirow{2}{*}{TRS}} & \multicolumn{1}{l|}{\multirow{2}{*}{BSS}} & \multicolumn{1}{c|}{\multirow{2}{*}{RA}} & \multicolumn{2}{c|}{Total-Text}  & \multicolumn{3}{c|}{ICDAR2015}  & \multicolumn{1}{c}{\multirow{2}{*}{Parameters}}    \\ \cline{5-9} 
   & \multicolumn{1}{c|}{}        & \multicolumn{1}{c|}{}                                     & \multicolumn{1}{c|}{}                                     & \multicolumn{1}{c|}{Det-Hmean} & \multicolumn{1}{c|}{E2E-Hmean} & \multicolumn{1}{c|}{S} & \multicolumn{1}{c|}{W} & \multicolumn{1}{c|}{G} & \multicolumn{1}{c}{}\\ \hline
$Baseline$  (Huang et al.,2022)  &  &  &  & \multicolumn{1}{c|}{83.2}      & \multicolumn{1}{c|}{66.9}    & \multicolumn{1}{c|}{79.6} & \multicolumn{1}{c|}{72.5} &  \multicolumn{1}{c|}{65.7}   &  \multicolumn{1}{c}{113.7M}    \\ \cline{1-1} \hline
$Baseline+$        &  \checkmark       &                                                                             &                                                        & \multicolumn{1}{c|}{80.2}      & 63.3     & \multicolumn{1}{c|}{76.0} & \multicolumn{1}{c|}{69.1} &  \multicolumn{1}{c|}{61.4}       &  \multicolumn{1}{c}{64.8M}           \\ \cline{1-1}
$Baseline+$         &  \checkmark   & \checkmark                                                                   &                                                          & \multicolumn{1}{c|}{83.6}      & 67.9       & \multicolumn{1}{c|}{79.5} & \multicolumn{1}{c|}{72.7} &  \multicolumn{1}{c|}{65.4}      &  \multicolumn{1}{c}{65.4M}              \\ \cline{1-1}
$Baseline+$         &  \checkmark   &                                                                   &  \checkmark                                                         & \multicolumn{1}{c|}{80.4}      & 65.1       & \multicolumn{1}{c|}{78.9} & \multicolumn{1}{c|}{71.3} &  \multicolumn{1}{c|}{64.7}      &  \multicolumn{1}{c}{70.0M}              \\ \cline{1-1}
\cline{1-1}
SwinTextSpotter v2   &  \checkmark      & \checkmark                                                             & \checkmark                                                         & \multicolumn{1}{c|}{\textbf{83.6}}      & \textbf{71.3}      & \multicolumn{1}{c|}{\textbf{81.8}} & \multicolumn{1}{c|}{\textbf{75.4}}  & \multicolumn{1}{c|}{\textbf{68.3}}      &  \multicolumn{1}{c}{70.6M}          \\
\cline{1-1} \hline

\end{tabular}}
\label{Ablation}
\end{table*}

\begin{table*}[!th]
\centering
\caption{Ablation studies on different refinement stages in the detector. The parameters contain parameters of the detector and recognizer.}
\begin{tabular}{@{}l|cc|ccc|c@{}}
\hline
\multirow{2}{*}{Refinement Stages} & \multicolumn{2}{c|}{Total-Text}  & \multicolumn{3}{c|}{ICDAR2015}  & \multicolumn{1}{c}{\multirow{2}{*}{Parameters}}    \\ \cline{2-6} 
  & \multicolumn{1}{c|}{Det-Hmean} & \multicolumn{1}{c|}{E2E-Hmean} & \multicolumn{1}{c|}{S} & \multicolumn{1}{c|}{W} & \multicolumn{1}{c|}{G} & \multicolumn{1}{c}{}\\ \hline
Two    & \multicolumn{1}{c|}{82.9}      & \multicolumn{1}{c|}{65.9}    & \multicolumn{1}{c|}{81.2} & \multicolumn{1}{c|}{74.1} &  \multicolumn{1}{c|}{67.0}   &  \multicolumn{1}{c}{57.2M}    \\ 
Three                                                        & \multicolumn{1}{c|}{83.6}      & 71.3      & \multicolumn{1}{c|}{81.8} & \multicolumn{1}{c|}{75.4}  & \multicolumn{1}{c|}{68.3}      &  \multicolumn{1}{c}{70.6M}           \\ 
Four        & \multicolumn{1}{c|}{84.0}      & 71.8       & \multicolumn{1}{c|}{82.2} & \multicolumn{1}{c|}{75.5} &  \multicolumn{1}{c|}{68.7}      &  \multicolumn{1}{c}{84.0M}              \\ 
Six       & \multicolumn{1}{c|}{\textbf{84.5}}      & \textbf{72.2}      & \multicolumn{1}{c|}{\textbf{82.6}} & \multicolumn{1}{c|}{\textbf{75.7}}  & \multicolumn{1}{c|}{\textbf{69.1}}      &  \multicolumn{1}{c}{97.4M}          \\
\cline{1-1} \hline
\end{tabular}
\label{ablation_num_stage}
\end{table*}

\begin{table*}[!h]
\centering
\caption{Ablation studies on Recognition Conversion and TLSAM without finetuning.}
\resizebox{0.7\linewidth}{!}{
\begin{tabular}{@{}l|cc|cc|ccc}
\hline
\multirow{2}{*}{Methods} & \multirow{2}{*}{RC} & \multirow{2}{*}{TLSAM} & \multicolumn{2}{c|}{Total-Text}  & \multicolumn{3}{c}{ICDAR2015}     \\ \cline{4-8} 
 & & & \multicolumn{1}{c|}{Det-Hmean} & \multicolumn{1}{c|}{E2E-Hmean} & \multicolumn{1}{c|}{S} & \multicolumn{1}{c|}{W} & \multicolumn{1}{c}{G}\\ \hline
SwinTextSpotter v2  & \checkmark & \checkmark & \multicolumn{1}{c|}{\textbf{83.6}}      & \textbf{71.3}      & \multicolumn{1}{c|}{\textbf{81.8}} & \multicolumn{1}{c|}{\textbf{75.4}}  & \multicolumn{1}{c}{\textbf{68.3}}          \\
SwinTextSpotter v2                                                         & $\times$ & \checkmark & \multicolumn{1}{c|}{81.7}    & 68.1      & \multicolumn{1}{c|}{79.3} & \multicolumn{1}{c|}{73.8}  & \multicolumn{1}{c}{65.0}      \\ 
SwinTextSpotter v2  & \checkmark & $\times$   & \multicolumn{1}{c|}{83.1}    & 70.6      & \multicolumn{1}{c|}{80.5} & \multicolumn{1}{c|}{74.3}  & \multicolumn{1}{c}{66.9}   \\ 
SwinTextSpotter v2  &  $\times$ & $\times$   & \multicolumn{1}{c|}{80.8}      & 66.4      & \multicolumn{1}{c|}{78.8} & \multicolumn{1}{c|}{73.2}  & \multicolumn{1}{c}{64.5}  \\
\cline{1-1} \hline
\end{tabular}}
\label{ablation_rc_tlsam}
\end{table*}

\begin{table}[t]
\centering
\caption{Effectiveness of two attention alignments without finetuning. All results are the end-to-end text spotting results on Total-Text. LA means the local-level alignment. GA means the global-level alignment.}
\begin{tabular}{c|c|c|c}
\hline
\multirow{2}*{LA} & \multirow{2}*{GA} & \multicolumn{2}{c}{Total-Text}  \\ \cline{3-4} 
           &      & None   & Full \\ \hline
 $\times$  & $\times$ &  67.9          & 77.0     \\ 
 \checkmark  & $\times$ &  69.4          & 81.2     \\ 
 $\times$   & \checkmark  & 68.7   & \multicolumn{1}{c}{80.4}  \\ 
 \checkmark    & \checkmark   & \textbf{71.3} & \textbf{81.7}   \\ 
\hline
\end{tabular}
\label{tab:two_level_align}
\end{table}

\begin{table}[!h]
\centering
\caption{Ablation study of different resolutions on ICDAR 2015. “S”, “W”, and “G” represent recognition with “Strong”, “Weak”, and “Generic” lexicon, respectively.}
\resizebox{\linewidth}{!}{%
\begin{tabular}{ll|ccc}
\hline
\multirow{2}{*}{Method} & \multirow{2}*{resolution} & \multicolumn{3}{c}{ICDAR 2015 End-to-End}                    \\ \cline{3-5} 
                    &    & \multicolumn{1}{c|}{S}    & \multicolumn{1}{c|}{W}    & G    \\ \hline
SwinTextSpotter v2   & 1600      & \multicolumn{1}{c|}{89.6} & \multicolumn{1}{c|}{84.1} & 79.4 \\ 
SwinTextSpotter v2   & 4000      & \multicolumn{1}{c|}{\textbf{90.0}} & \multicolumn{1}{c|}{\textbf{84.5}} & \textbf{80.1} \\ \hline
\end{tabular}}
\label{icdar2015_resolution}
\end{table}

\begin{table}[t]
\centering
\caption{Effectiveness of adding Recognition Alignment on previous text spotting methods. All results are the end-to-end text spotting results on Total-Text. RA means the Recognition Alignment.}
\resizebox{0.7\linewidth}{!}{
\begin{tabular}{l|c|c|c}
\hline
\multirow{2}*{Method} & \multirow{2}*{RA} & \multicolumn{2}{c}{Total-Text}  \\ \cline{3-4} 
           &      & None   & Full \\ \hline
Mask TextSpotter v3     & $\times$ &  75.1          & 81.4     \\ 
Mask TextSpotter v3    & \checkmark  & 76.3   & \multicolumn{1}{c}{82.3}  \\ 
ABCNet v2        & $\times$   & 70.4      &78.1       \\ 
ABCNet v2        & \checkmark    & 71.2    & 82.6 \\ 
\hline
\end{tabular}}
\label{tab:samplepoint_previous}
\end{table}

\subsection{Ablation Studies}
To investigate the effectiveness of the proposed components, we perform ablation studies on the Total-Text dataset and the ICDAR2015 dataset. In this section, all experiments are conducted by training the model on the Curved SynthText, ICDAR-MLT, and the corresponding datasets without fine-tuning, with the exception of the ablation study of different resolutions on ICDAR2015.

\textbf{Three Refinement Stages (TRS)}. To decrease model complexity and the parameters, we make modifications to the detector architecture. We reduce the number of refinement stages from six to three and remove the use of dilated convolutions in comparison to the conference version. Table~\ref{Ablation} presents the results of our modifications. We observe a decrease in performance when the refinement stages were reduced. Specifically, on Total-text, we note a decrease of 3$\%$ and 3.3$\%$ in text detection and spotting results, respectively. Additionally, on ICDAR2015, we observe a performance decrease of 4.3$\%$ on the `General` lexicon. Moreover, as shown in Table \ref{Ablation}, the number of parameters in the detector is reduced by approximately two times. Specifically, the parameters of the detector and recognizer are reduced from 113.7M to 64.8M.

\begin{figure}[h]
    \centering
    \begin{subfigure}{0.25\textwidth}
        \centering
        \includegraphics[width=1.0\linewidth]{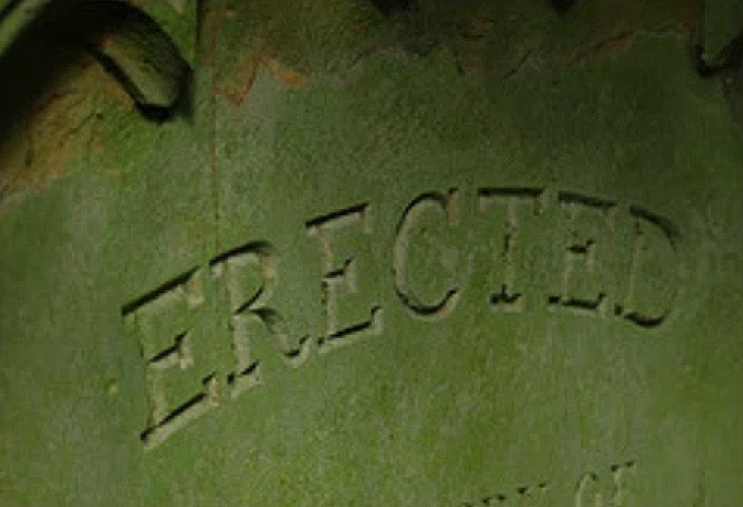} 
        \caption{}
        \label{fig:sub1}
    \end{subfigure}%
    ~
    \begin{subfigure}{0.25\textwidth}
        \centering
        \includegraphics[width=1.0\linewidth]{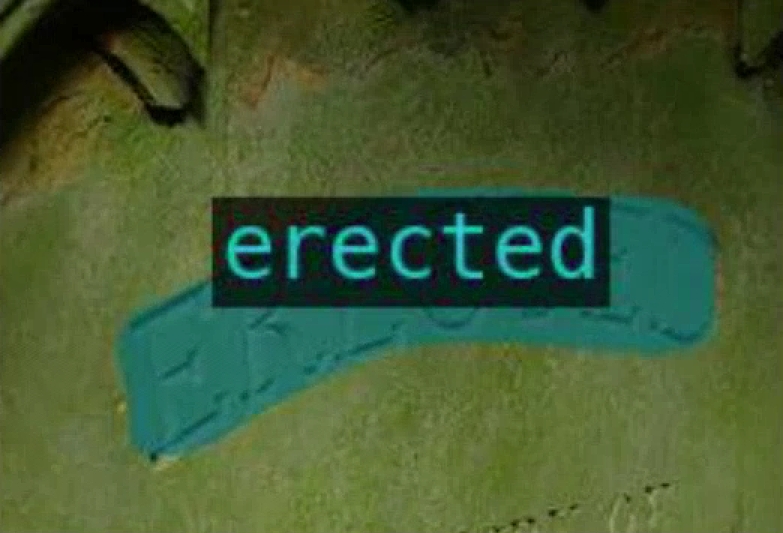} 
        \caption{}
        \label{fig:sub3}
    \end{subfigure}
    \caption{(a) represents the input image. (b) represents the prediction of SwinTextSpotterv2
}
    \label{fig:obscured}
\end{figure}

\textbf{Box Selection Schedule (BSS)}. To demonstrate the effectiveness of the Box Selection Schedule (BSS), we conduct experiments on the \textbf{TRS}. We present the results of the TRS equipped with the BSS in the fourth row of Table~\ref{Ablation}. We observe that the BSS significantly enhanced the performance of our model. On Total-Text, the text detection accuracy is improved by 3.4$\%$, and the end-to-end text spotting result is improved by 4.6$\%$. Additionally, on ICDAR2015 for the `General' lexicon, the end-to-end text spotting result is improved by 4$\%$. By generating high-quality proposal boxes, the BSS not only improves the precision of the text localization but also enhances the overall robustness of the system. Additionally, these enhancements come with only a minor increase in the number of parameters, approximately 0.6M. The box Selection schedule will first generate a set of high-recall proposals. This ensures that even challenging text instances, such as those that are partially obscured or intermingled with complex backgrounds, are included in the initial set of proposals. Then, the Top-K selection process identifies and selects all regions that potentially contain text. Even if the initial proposals are not very accurate, the detector's refinement process will further adjust the proposals' positions to achieve more precise detection results. Fig.~\ref{fig:obscured} provides a visualization result of a challenging case. It can be found that our method can effectively handle text that may be partially obscured or intermingled with complex backgrounds.

\textbf{Recognition Alignment (RA)}. To demonstrate the effectiveness of the Recognition Alignment (RA), we integrated it into the baseline+TRS model. As presented in the third row of Table \ref{Ablation}, we find that incorporating RA into the baseline+TRS model enhances the end-to-end text spotting performance. Specifically, we observe an improvement of 1.8\% on the TotalText dataset and 3.3\% on the `General' metric of ICDAR2015. The integration of RA aligns the recognition feature representations more closely with the textual content, thus facilitating a more accurate recognition process and leading to overall performance gains. Since RA is designed to dynamically adapt recognition features, its primary benefit is on the recognition, with only a minor effect on detection performance.

Moreover, the integration of RA, TRS, and BSS further enhances text spotting performance, as shown in the fifth row of Table \ref{Ablation}. Notably, there is a 3.4\% boost in performance on the Total-Text dataset and a 2.9\% improvement on the `General' lexicon of ICDAR2015. These results demonstrate the effectiveness of the RA in improving text spotting performance.

In addition, we also add Recognition Alignment to previously proposed methods~\citep{liao2020mask,liu2021abcnetv2}. The results, as displayed in Table~\ref{tab:samplepoint_previous}, indicate that Recognition Alignment has the potential to enhance the performance of existing methods. 

\textbf{Refinement Stages}. We also conduct experiments to evaluate the impact of various refinement stages. The results are presented in Table~\ref{ablation_num_stage}. Specifically, we observe that increasing the number of refinement stages from two to three led to a significant enhancement in end-to-end text spotting performance for both the TotalText and ICDAR2015 datasets. For the TotalText dataset, this adjustment result in a 5.4\% increase in accuracy, while for the `General' lexicon of ICDAR2015, there is a 1.3\% improvement. Gradually increasing the number of refinement stages, the performance continues to improve, but the improvement is relatively smaller compared to the increase from two to three stages. Considering the balance between performance improvements and model complexity, we choose three refinement stages as our default setting.

\textbf{Two-Level Attention Alignments}. We conduct experiments to evaluate the contributions of two types of attention alignments to overall model performance, with the results presented in Table \ref{tab:two_level_align}. Our findings indicate that both types of attention alignment can enhance the model's performance. Specifically, the local-level attention alignment aligns text details locally, while the global-level attention alignment aligns global semantic features. The performance improvement achieved by the local-level attention alignment surpasses that of the global-level attention alignment, suggesting that detailed local text features have a greater influence on recognition tasks. Furthermore, combining both local and global attention alignments brings even more improvement, achieving a 3.4\% gain on the `None' metric of TotalText, which demonstrates the effectiveness of our proposed approach.

\textbf{Recognition Conversion}. As shown in Table \ref{ablation_rc_tlsam}, removing the RC module leads to a performance degradation of 1.9\% and 3.2\% in detection and text spotting tasks on the TotalText dataset, respectively. Furthermore, the removal of the RC module also causes a decline in performance on the ICDAR2015 dataset: 2.5\%, 1.6\%, and 3.3\% across the `Strong', `Weak', and `Generic' lexicons, respectively. These results indicate that RC improves the performance of the detector and recognizer. This is mainly because the RC enhances the synergy between detection and recognition by back-propagating recognition information to the detector and adapting the detection features to assist recognition.

\textbf{Two-level self-attention mechanisms}. 
In Table \ref{ablation_rc_tlsam}, we further conduct experiments to explore the influence of TLSAM. Removing the TLSAM module results in a performance degradation of 1.3\%, 1.1\%, and 1.4\% for the `Strong', `Weak', and `Generic' lexicons on the ICDAR2015 dataset, respectively. TLSAM provides fine-grained features that enhance recognition. Simultaneously removing both RC and TLSAM leads to significant drops in performance by 2.8\% on detection and 4.9\% on text spotting. These results demonstrate the effectiveness of TLSAM.

\textbf{Scalability}. Inspired by~\citep{ye2023deepsolo}, we extend our methodology to incorporate the large-scale TextOCR dataset~\citep{singh2021textocr} to evaluate the scalability of our proposed method. Our experiments, summarized in Tables \ref{Rotate-IC13-end-to-end}, \ref{VinText}, \ref{tab:ctw1500_e2e}, \ref{ICDAR 2015 End-to-End recognition result}, \ref{tab:totaltext_e2e}, and \ref{tab:inverse_e2e}, reveal a significant performance enhancement on various benchmarks when leveraging a larger training dataset. This improvement demonstrates that by enhancing the synergy between detection and recognition, our method achieves robust scalability.

\textbf{High-Resolution}. In Table \ref{icdar2015_resolution}, we conduct experiments to demonstrate our method's ability to handle high-resolution images. The results indicate that increasing the resolution from 1600 to 4000 yields a 0.4\%, 0.4\%, and 0.7\% improvement in the `Strong', `Weak', and `Generic' lexicons of ICDAR2015, respectively. This demonstrates the scalability of our model across different resolutions.

\subsection{Limitations}
\label{subsec:discuss}
The proposed SwinTextSpotter v2 achieves state-of-the-art performance on several scene text benchmarks. However, SwinTextSpotter v2 still faces challenges in handling specific situations, including horizontally arranged vertical texts and artistic texts with colors and styles that closely resemble the background. 

We visualize these typical errors in Fig.~\ref{fig:vis_results} (h). As shown on the left of the figure, we can observe that the “attccs” is mistakenly recognized as “ffood”. This is because horizontally arranged vertical texts are rare in the datasets and most text is arranged from left to right. For the right of the figure, the text ``g'' almost blends into the background, which makes it difficult for the detector to detect the text. Previous methods~\citep{liu2021abcnetv2,zhang2022text,fangabinet++} also struggle with such specific situations and there is room for further research on how to effectively extract text content in these cases. For the inference speed, SwinTextSpotter v2 can run at $4.0$ FPS which is comparable to another state-of-the-art model GLASS~\citep{ronen2022glass} ($3.0$ FPS).

\section{Conclusion}
In this study, we make a brand-new attempt that delves deeper into the interplay between the detector and the recognizer. 
Using \textit{Recognition Conversion}, SwinTextSpotter v2 can effectively suppress background noise in the recognition process by making detection results differentiable with respect to recognition loss. Additionally, \textit{Recognition Alignment} allows the recognizer to dynamically choose features, addressing the misalignment issue in text spotting. By integrating these improvements, SwinTextSpotter v2 eliminates the need for a rectification module and optimizes both detection and recognition modules for enhanced spotting performance.
Furthermore, we adopt a Box Selection Schedule that significantly reduces the detector parameters while achieving better performance. Extensive experiments on public benchmarks demonstrate that the success of end-to-end text spotting relies on the synergistic relationship between the two tasks. We hope that our work could inspire further development of synergistic techniques in the future.

\section*{Acknowledgement}
This research is supported in part by National Natural Science
 Foundation of China (Grant No.: 62206104, 62476093, 62225603), National Key R\&D Program of China (Grant No.: 2022YFC2305102).

\bibliographystyle{spbasic}     
\bibliography{main}

\end{document}